\documentclass[10pt,twocolumn,letterpaper]{article}

\usepackage[pagenumbers]{cvpr} 

\usepackage{multirow}
\usepackage{tabularx}
\usepackage{caption}
\usepackage{subcaption}
\usepackage{adjustbox}
\usepackage{soul}
\usepackage{wrapfig}
\usepackage{epsfig}
\usepackage{graphicx}
\usepackage{array}        
\usepackage{booktabs}     
\newcommand{\textoverline}[1]{$\overline{\mbox{#1}}$}

\usepackage{lipsum}
\newcommand\blfootnote[1]{%
  \begingroup
  \renewcommand\thefootnote{}\footnote{#1}%
  \addtocounter{footnote}{-1}%
  \endgroup
}

\definecolor{cvprblue}{rgb}{0.21,0.49,0.74}
\usepackage[pagebackref,breaklinks,colorlinks,allcolors=cvprblue]{hyperref}

\def\paperID{9} 
\def\confName{CVPR}
\def\confYear{2025}

\title{Understanding Depth and Height Perception in Large Visual-Language Models}

\author{Shehreen Azad\textsuperscript{1*} \qquad  Yash Jain \textsuperscript{2} \qquad Rishit Garg\textsuperscript{3} \qquad
Yogesh S Rawat\textsuperscript{1} \qquad Vibhav Vineet\textsuperscript{2} \\
\vspace{1pt}\\
\textsuperscript{1}Center for Research in Computer Vision, University of Central Florida;\\
\textsuperscript{2}Microsoft Research; \qquad \textsuperscript{3}Indian Institute of Technology, Kharagpur. 
\\ 
\textit{\href{https://sacrcv.github.io/GeoMeter-website/}{Project Page} \qquad \href{https://github.com/sacrcv/GeoMeter}{Code}}
}
\begin{document}

\twocolumn[{%
\renewcommand\twocolumn[1][]{#1}%
\maketitle
\vspace{-10mm}
\begin{center}
    \centering
    \captionsetup{type=figure}
    \includegraphics[width=\linewidth]{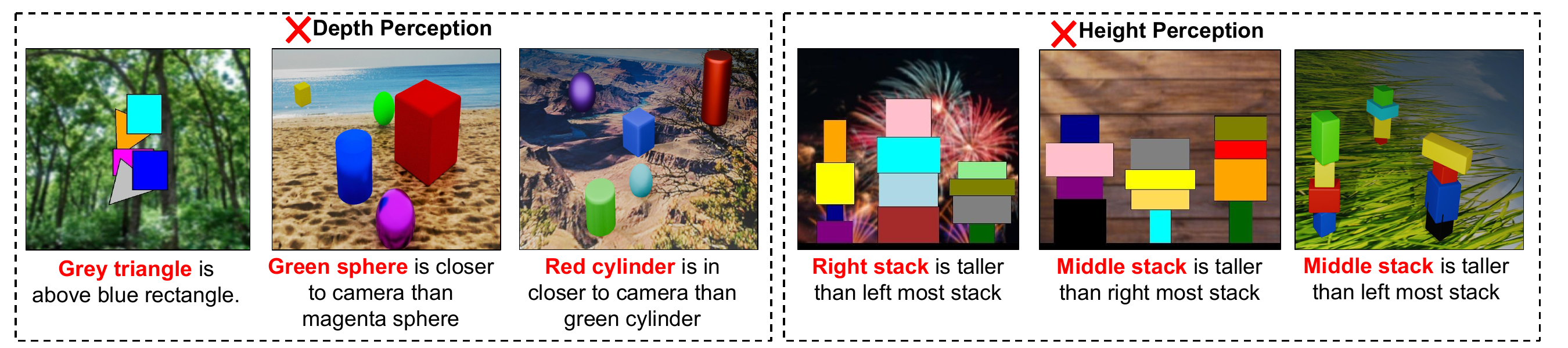}
    \captionof{figure}{\textbf{Depth and height perception capability of existing VLM.} Here, we show failure cases of GPT-4V in understanding depth and height on GeoMeter, our proposed suite of benchmark datasets.}
    \label{fig:teaser}
\end{center}%
}]

\begin{abstract}
Geometric understanding—including depth and height perception—is fundamental to intelligence and crucial for navigating our environment. Despite the impressive capabilities of large Vision Language Models (VLMs), it remains unclear how well they possess the geometric understanding required for practical applications in visual perception. In this work, we focus on evaluating the geometric understanding of these models, specifically targeting their ability to perceive the depth and height of objects in an image. To address this, we introduce GeoMeter, a suite of benchmark datasets—encompassing 2D and 3D scenarios—to rigorously evaluate these aspects. By benchmarking 18 state-of-the-art VLMs, we found that although they excel in perceiving basic geometric properties like shape and size, they consistently struggle with depth and height perception. Our analysis reveal that these challenges stem from shortcomings in their depth and height reasoning capabilities and inherent biases. This study aims to pave the way for developing VLMs with enhanced geometric understanding by emphasizing depth and height perception as critical components necessary for real-world applications. 
\end{abstract}
\vspace{-7mm}
\section{Introduction}
\label{sec:intro}

In recent years, the AI community has significantly focused on integrating visual and natural language inputs, notably in Visual Question Answering (VQA) systems. These systems 
\blfootnote{\textsuperscript{*}Corresponding author. Shehreen.Azad@ucf.edu}
analyze images and answer questions about them, showing substantial advancements in understanding basic visual concepts such as shape identification ~\citep{kuhnle2017shapeworld}, object detection ~\citep{zou2023object}, and the spatial relationships ~\citep{johnson2017clevr, chen2024spatialvlm, liu2023visual} by using large Visual Language Models (VLMs). 
These models have excelled in processing complex text and visual inputs due to their strong visual understanding capability, leading to applications in image captioning, visual question answering, image text retrieval, and so on. 

The ability to understand visual properties such as size, shape, depth, and height is fundamental to visual understanding, yet many existing Visual Question Answering (VQA) benchmarks ~\citep{johnson2017clevr, chen2024spatialvlm, liu2023visual, diwan2022winoground, thrush2022winoground} do not specifically focus on the depth and height perception capabilities of Vision Language Models (VLMs). Accurate perception of these dimensions is vital for practical applications like surveillance, navigation, and assistive technologies. The lack of accurate depth and height understanding in VLMs can lead to serious consequences, such as misjudging the proximity of objects, which could result in catastrophic outcomes in real-world scenarios. 

Despite VLMs' abilities to recognize object shapes and sizes, their depth and height reasoning often relies on learned size/shape cues rather than actual spatial analysis, potentially influenced by biases from training data ~\citep{jayaraman2024d}. Alternatively, models might estimate the depth based on the apparent size of objects, without genuine inter-object reasoning. 
An example illustrated in Figure \ref{fig:teaser} shows how GPT-4V ~\citep{openai2024gpt4}, one of the most popular closed-source VLMs, struggles with depth and height perception in images featuring basic 2D or 3D objects. The model is just randomly answering without correctly assessing the spatial relationship between the objects, relying on visual cues that conflict with their actual arrangement. Additional examples in Figure \ref{fig:teaser} further demonstrate GPT-4V's failures in perceiving depth and height. These limitations highlight the need to explore such shortcomings more thoroughly and develop targeted benchmarks and training strategies that can better equip VLMs to handle complex, real-world environments with accurate depth and height perception.

In this paper, we aim to evaluate the depth and height reasoning capabilities of Vision Language Models (VLMs) to identify their strengths and limitations in visual perception. While auxiliary sensors play a crucial role in depth estimation and other alternative methods of estimating depth and height may outperform visual language models (VLMs) in specific tasks, our research aims to assess the standalone capabilities of VLMs rather than suggesting their replacement. To achieve this, we design GeoMeter, a suite of synthetic benchmark datasets focusing on 2D and 3D scenarios, named GeoMeter-2D and GeoMeter-3D respectively. These probing datasets, feature basic shapes, such as rectangles, circles, cubes, and cylinders, and are crafted to test the visual reasoning capabilities of VLMs. The development of synthetic datasets is motivated by concerns about test-time data leakage, where large VLMs, trained on vast datasets, might encounter images during testing that they have already seen during training. We prioritize clean, programmatically generated data over mere size to ensure that the evaluation is not compromised by dataset familiarity. This controlled approach minimizes the risk of data leakage and enables a more focused and precise assessment of VLMs’ understanding of depth and height, free from the confounding influence of real-world cues present in many publicly sourced datasets. To this end, our probing datasets consist of around $4$k unique images and $11.2$k image text pairs, designed to probe depth and height reasoning in VLMs.

We extensively analyze our proposed suite of benchmark datasets on \textit{$18$} recent open-source and closed-source models for the VQA task. Our findings reveal several key insights: (1) While VLMs demonstrate basic geometric understanding, they struggle significantly with depth and height perception tasks. 
(2) Models generally show better depth perception than height, likely due to the more common and simpler depth cues like occlusion and perspective, which are prevalent in training datasets, making depth easier to process than the more complex cues required for accurate height estimation. (3) The lack of depth and height perception ability stems from the models' intrinsic visual reasoning abilities rather than the level of prompt detail. (4) Inherent biases are evident in models' responses when faced with advanced perception tasks.

Overall, our contributions can be summarized as follows:
\begin{itemize} 
    \item We examine the depth and height reasoning abilities of VLMs, identifying strengths, limitations, and areas for improvement in visual perception tasks.
    \item Our analysis spans $18$ VLMs, both open-source and closed-source, revealing behavioral patterns and biases in depth and height perception.
    \item To support this evaluation, we introduce GeoMeter, comprising two datasets - GeoMeter-2D and GeoMeter-3D - that test VLMs on depth and height perception tasks.

\end{itemize}
\section{Related Works}
\label{sec:rel_works}
\textbf{Visual Language Models (VLMs).}
The field of AI has undergone a significant transformation with the advent of vision language models (VLMs), which are trained on extensive multimodal datasets and are versatile across numerous applications ~\citep{radford2021learning, liu2023llava, li2023blip, dai2024instructblip, sun2023eva, li2025videomamba, sun2019videobert, azad2025hierarq}. These models have shown remarkable performance in language and vision-related tasks, e.g. recognition, reasoning, etc. VLMs are models with a pre-trained LLM backbone and a vision encoder; which are aligned by using different methods. Recent closed-source VLMs such as GPT-4 ~\citep{openai2024gpt4}, Gemini ~\citep{geminiteam2024gemini}, Claude ~\citep{claude} showcase a strong potential for tasks that require understanding and processing information across different modalities. Additionally, various openly available VLMs such as LLaVA ~\citep{liu2023llava}, LLaVA-NeXT ~\citep{liu2023improvedllava}, Bunny ~\citep{he2024bunny} etc. also have comparative performance with the closed-source models across different vision-language tasks. All of these VLMs are trained on massive amount of public and proprietary data, making them a strong performer of general reasoning.

\begin{table}[t!]
    \centering
    \caption{\textbf{Dataset statistics} of our proposed benchmark suites. 
    }
    \resizebox{0.99\linewidth}{!}{
    \begin{tabular}{c|m{1.4cm}|m{3.6cm}|c|m{1.2cm}}
        \hline
         Dataset & Perception Task & Description& Images & Img-Text pairs\\
         \hline
         \multirow[c]{4}{1.3cm}{GeoMeter-2D} & Depth  & Determine which object is on the top. &  $1200$& \multirow[c]{4}{0.5cm}{$4800$}\\
         \cline{2-4}
         & Height &  Provide height ordering from shortest to tallest among given stacks& $1200$ & \\
         \hline
         \multirow[c]{4}{1.3cm}{GeoMeter-3D} & Depth & Determine which object is closer to the camera. &  $800$& \multirow[c]{4}{0.5cm}{$6400$}\\
         \cline{2-4}
         & Height &  Provide height ordering from shortest to tallest among given stacks& $800$ & \\
         \hline
    \end{tabular}}
    \label{tab:stat}
    \vspace{-4mm}
\end{table}

\noindent
\textbf{Exploring Visual Reasoning Capability of VLMs.} Previous works have extensively explored the spatial reasoning and object understanding capabilities of Vision Language Models (VLMs), probing their ability to grasp object-attribute relationships and spatial concepts like spatial reasoning through various benchmarks ~\citep{thrush2022winoground, diwan2022winoground, johnson2017clevr, krishna2017visual, liu2023visual, schiappa2024probing, schiappa2024robustness, huang2024a3vlm, tong2024cambrian}. However, specific geometric properties such as depth and height perception have been largely under-explored. While there are benchmarks that assess geometric property understanding ~\citep{chen2021geoqa,zhang2024geoeval,sun2024advancing}, they often rely on mathematical knowledge and do not directly probe these properties in the context of natural visual understanding.
Moreover, many of the datasets used in these studies ~\citep{thrush2022winoground, diwan2022winoground, krishna2017visual, liu2023visual, schiappa2024probing, tong2024cambrian} are curated from pre-existing datasets and/or the internet, which introduces the risk of data leakage during testing, making it difficult to assess VLMs' true capability for depth and height reasoning. Although synthetic datasets have been developed ~\citep{johnson2017clevr, kuhnle2017shapeworld}; they are not specifically tailored to tasks focusing on depth and height understanding, further limiting their effectiveness in thoroughly evaluating these advanced visual concepts. Our proposed benchmark suite addresses this gap by offering image-text pairs that target depth and height perception, without relying on mathematical reasoning, providing a more focused assessment of VLMs in this area.

\section{Benchmark}
\label{sec:benchmark_and_eval}
\begin{figure}[t!]
    \centering
    \includegraphics[width=\linewidth]{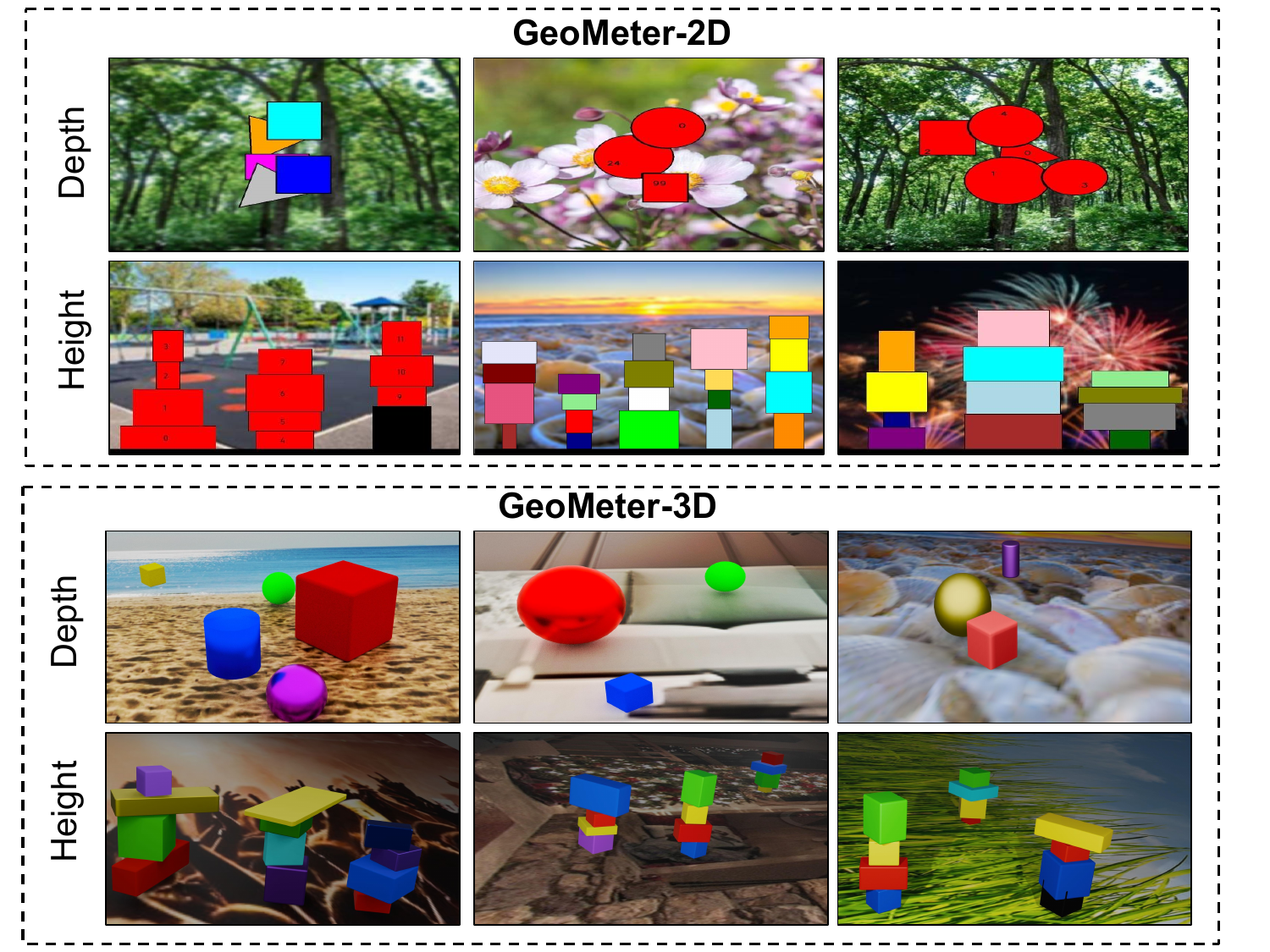} 
    \caption{\textbf{Samples from the proposed suite of benchmark datasets.} Here each samples are shown with random query attributes- color and numeric label for GeoMeter-2D; and color and material for GeoMeter-3D dataset.}
    \vspace{-10pt}
    \label{fig:samples}
\end{figure}

\vspace{-2mm}
Our proposed suite of benchmark datasets consist of GeoMeter-2D, and GeoMeter-3D datasets that are designed to test model performance on depth and height perception tasks, utilizing unique identifiers as diverse query attributes for question generation. Table \ref{tab:stat}, and Figure \ref{fig:samples} respectively show the dataset statistics and sample images of our proposed datasets. More samples from each dataset is given in the appendix. In the following sections, we describe the detailed data generation process for our benchmark.

\subsection{Datasets}
\label{subsec:datasets}
The generation process is divided into Image generation (Section \ref{subsubsec:img}) and Question generation (Section \ref{subsubsec:ques}).

\subsubsection{Image Generation}
\label{subsubsec:img}
Our proposed synthetic datasets are divided into two categories - \textit{Depth} and \textit{Height}, with each image containing a real-world scene background to add realism while maintaining controlled, programmatically generated content. We generate images in two variety of scene density - 3 shapes and 5 shapes, with each shape having one unique identifier which is used as query attribute to refer to that certain object while probing the VLMs' depth and height perception.

\noindent
\textbf{GeoMeter-2D} dataset includes \textit{$2400$ images and $4800$ unique questions}, and is designed to test depth and height perception through basic 2D shapes. The \textit{Depth} category features overlapping $3$ or $5$ geometric shapes, like rectangles, triangles, and circles, positioned to create depth illusions. Ground truth for these images is stored in a scene graph that annotates each object’s shape, size, color, and spatial positioning, including depth ordering through directed edges connecting overlapping objects. Each object is assigned a unique identifier based on color and numeric labels.
For the \textit{Height} category, we generated scenes featuring sequentially labeled $3$ or $5$ towers, each consisting of four stacked rectangles. Each tower was created by randomizing the height and width of the individual rectangles to add variability to the scene. The bottom-most rectangle in some images is placed on a black strip representing an elevated platform, making the tower effectively shorter by one rectangle in actual height but visually elevated. These images are categorized into two subgroups: \textit{w/ step} for towers placed on a platform and \textit{w/o step} for towers directly placed on the ground. This setup allows VLMs to be rigorously tested on height comparison tasks, requiring them to correctly interpret both the visual cues of the towers’ absolute and relative heights, and the additional complexity introduced by the raised platforms. Each object in the scene is uniquely identified by its color and label, and the scene graph provides the ground truth, detailing the size, position, and elevation of each tower. 
 
\begin{figure*}[t!]
    \centering
    \includegraphics[width=\linewidth]{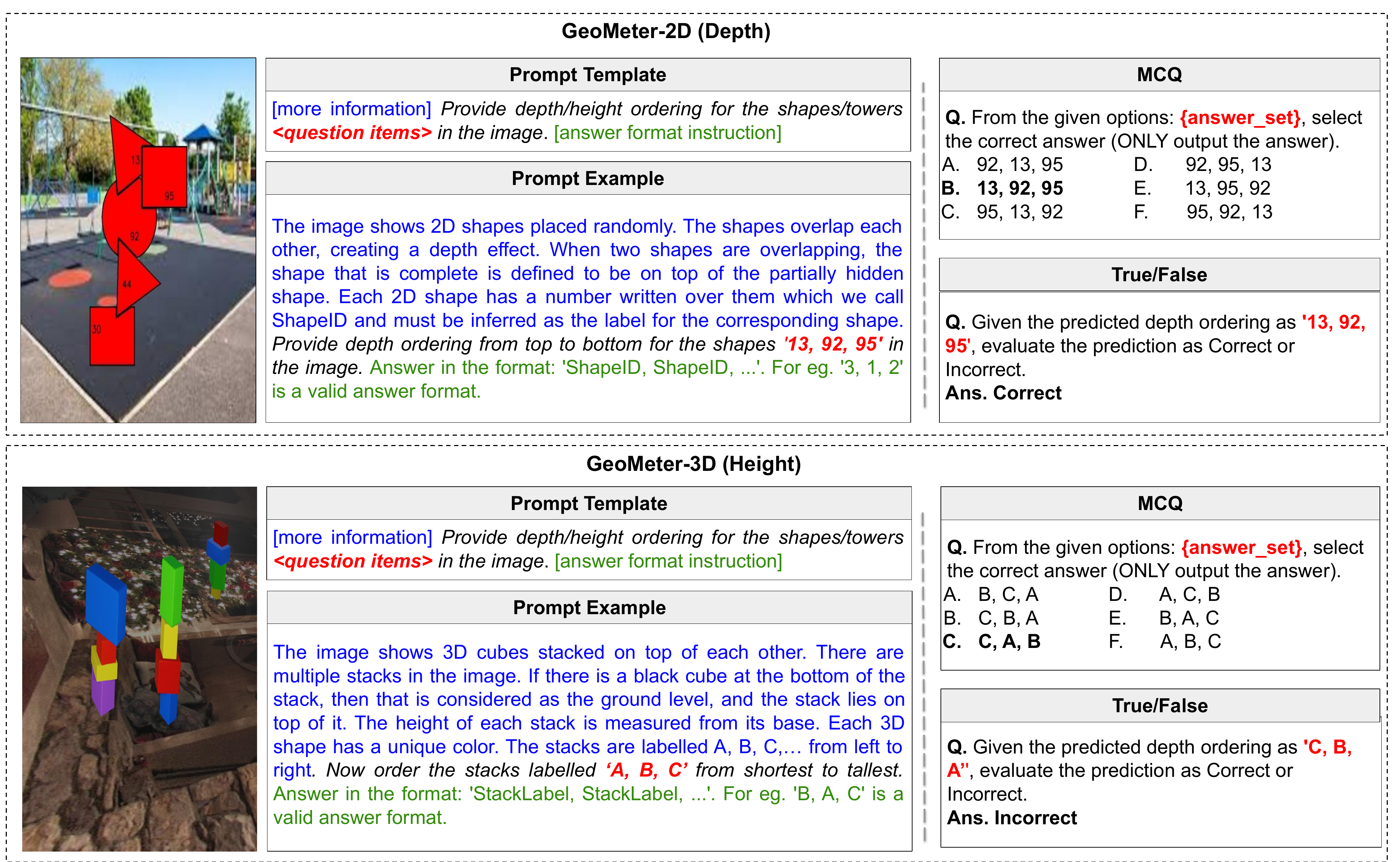}
    \caption{\textbf{Sample image-text pair from the datasets.} Here, prompt template shows the basic template for each image-text pair in our datasets, where the prompt example is the actual prompt for the image. The prompt example is appended with either MCQ or True/False type question.}
    \label{fig:prompt}
    \vspace{-4mm}
\end{figure*}
\noindent
\textbf{GeoMeter-3D} dataset consists of \textit{$1600$ images and $6400$ unique questions}, created based on the existing CLEVR dataset ~\citep{johnson2017clevr}. Scenes are generated using Blender ~\citep{blender}, with random jittering of light and camera positions to ensure variety. Objects in these scenes are annotated using a scene graph, which records each object’s shape, size, color, material (shiny ``metal" or matte ``rubber"), and position on the ground plane. The \textit{Depth} category includes randomly placed $3$ or $5$ cubes, spheres, and cylinders with distinct colors and materials as unique identifiers. These shapes are colored from a palette of eight colors and two materials, with increased horizontal and vertical margins than original CLEVR images between objects to reduce ambiguous spatial relationships. The scene graph captures all ground-truth information required to evaluate depth perception tasks, such as object distances and relative positions.
For the \textit{Height} category, same as the GeoMeter-2D dataset's height category setup, we created scenes with 3 or 5 towers, each consisting of four cubes stacked on top of each other. We created a base tower mesh and randomized each cube’s size, color, and material (either shiny ``metal" or matte ``rubber") for every image. Same as GeoMeter-2D, in some scenes, the bottom-most cube is black and matte, representing an elevated platform. The ground truth for these images is represented in the scene graph, detailing the exact size, position, and elevation of each tower. 

\subsubsection{Question Generation}
\label{subsubsec:ques}
The method used for generating questions is consistent across all our proposed datasets. Each question is a \textit{Description prompt} appended with an \textit{Answer format instruction}. The description prompt contains some general information about the scene providing semantic cues to the given image; followed by the actual question and answer format instruction. For example,
\textit{``[more information] Provide depth/height ordering for the shapes $<$question items$>$ in the image. [more information]"} is a descriptive prompt. This is followed by \textit{``From the given options: $<$answer set$>$, select the correct answer [more information]."} which is an answer format instruction.

The \textit{question items} is a list containing \textit{$<$query attribute$>$} appended by \textit{$<$shape$>$}. Here \textit{$<$query attributes$>$} is one of the unique identifiers of the dataset. 
For example in the question item \textit{``green metal cube"}, \textit{``green metal"} is the \textit{$<$query attribute$>$} and \textit{$<$cube$>$} is the shape.  The \textit{answer set} contains all possible valid values (\textit{$<$query attribute$>$} + \textit{$<$shape$>$}) to that given prompt. To generate both the question items and answer set, we read through the scene graph and run depth-first search on it to generate valid unambiguous values of object-pair relationship. For each image, there are two types of questions - MCQ and True/False. Some example prompts along with their corresponding image is shown in Figure \ref{fig:prompt}.

\section{Experimental Setup}
\label{sec:experiment}

\begin{table*}[t!]
    \centering
    \caption{\textbf{Performance comparison of the studied models on proposed datasets.} The reported results are averaged across depth and height category, query attributes and scene density with top scores in bold. Average denotes average performance of both datasets. Here, T/F denotes True/False type questions. }
    \resizebox{.85\linewidth}{!}{
\begin{tabular}{c|l|cc|cc|cc}
         \hline
          & \multirow{2}{*}{Model} & \multicolumn{2}{c|}{GeoMeter-2D} & \multicolumn{2}{c|}{GeoMeter-3D} & \multicolumn{2}{c}{Average} \\
          &  & MCQ          & T/F        & MCQ          & T/F & MCQ          & T/F        \\
          \hline
\multirow{14}{*}{\rotatebox[]{90}{Open}}  & LLaVA 1.5 7B \cite{liu2023improvedllava}                              & $\textbf{28.8}$        & $50.5$              & $28.0$        & $49.8$    & $28.4$   & $50.2$              \\
       & LLaVA 1.5 13B    \cite{liu2023improvedllava}                          & $17.8$        & $52.5$              & $29.0$        & $51.3$    & $23.4$   & $51.9$                  \\
       & LLaVA 1.6 Mistral 7B       \cite{liu2024llavanext}                & $22.1$         & $52.2$            & $26.7$        & $48.7$        & $24.4$   & $50.5$           \\
       & LLaVA 1.6 Vicuna 7B      \cite{liu2024llavanext}                  & $17.1$         & $51.7$              & $28.6$        & $50.0$     & $22.9$   & $50.9$               \\
       & LLaVA 1.6 Vicuna 13B    \cite{liu2024llavanext}                   & $28.2$        & $\textbf{54.2}$              & $32.5$       & $52.7$      & $30.4$   & $\textbf{53.5}$              \\
       & Bunny-v1.0-3B     \cite{he2024bunny}                         & $24.1$        & $50.1$             & $17.1$        & $37.1$     & $20.6$   & $43.6$                 \\
       & Bunny-v1.0-4B         \cite{he2024bunny}                     & $24.2$         & $52.6$              & $19.9$       & $39.3$   & $22.1$   & $46.0$                   \\
       & Bunny-v1.1-4B         \cite{he2024bunny}                     & $26.6$        & $52.3$             & $26.9$         & $44.4$    & $26.8$   & $48.4$                 \\
       & Bunny-Llama-3-8B-V     \cite{he2024bunny}                    & $27.9$           & $50.2$              & $26.9$        & $43.2$   & $27.4$   & $46.7$                \\
       & Fuyu-8B     \cite{fuyu-8b}                               & $8.6$       & $53.0$              & $19.4$        & $43.2$     & $14.0$   & $48.1$                 \\
       & InstructBLIP-Flan-T5-XL    \cite{dai2024instructblip}                & $10.8$         & $47.4$              & $37.5$        & $52.1$    & $24.2$   & $49.8$               \\
       & InstructBLIP-Vicuna-7B        \cite{dai2024instructblip}             & $28.3$         & $49.0$              & $38.1$        & $53.8$     & $\textbf{33.2}$   & $51.4$               \\
       & LLaMA-Adapter-v2-Multimodal \cite{gao2023llamaadapterv2}                & $22.9$         & $48.8$            & $32.7$         & $52.4$   & $27.8$   & $50.6$                \\
       & MiniGPT-4           \cite{zhu2023minigpt}                       & $25.0$       & $50.4$              & $\textbf{39.4}$         &$\textbf{56.3}$  & $32.2$   & $53.4$                   \\
       \hline
       \hline
\multirow{4}{*}{\rotatebox[]{90}{Closed}} & GPT-4V \cite{achiam2023gpt}                                    & $25.5$         &  $54.0$             & $35.2$         & $50.5$                 & $30.4$   & $52.3$\\
       & GPT-4o          \cite{achiam2023gpt}                           & $\textbf{30.8}$         & $\textbf{56.7}$     & $\textbf{38.5}$       & $\textbf{52.4}$              & $\textbf{34.7}$   & $\textbf{54.6}$     \\
       & Claude 3 Opus    \cite{claude}                          & $29.0$        & $51.9$              & $36.2$         & $49.9$    & $32.6$   & $50.9$               \\
       & Gemini 1.5 Pro \cite{geminiteam2024gemini} & $28.8$ & $54.5$ & $36.5$ & $51.0$ & $32.7$   & $52.8$ \\
       \hline
       \hline
       & Human evaluators & $\textbf{91.0}$ & $\textbf{99.0}$ & $\textbf{90.5}$ & $\textbf{97.0}$ & $\textbf{90.8}$ & $\textbf{98.0}$\\
       \hline
\end{tabular}
}
\label{tab:avg}
\end{table*}

\subsection{Vision Language Models}
\label{subsec:models}
We perform our benchmark evaluation on \textbf{$18$} state-of-the-art visual-language models. All of our chosen VLMs are trained on very large (public and/or proprietary) datasets. The selected VLMs can be categorized into \textbf{$14$} open-source and \textbf{$4$} closed-sourced models.

\noindent
\textbf{Open-source models.} We evaluate a range of open-source multimodal models: \textit{LLaVA} \cite{liu2023llava, liu2023improvedllava} and \textit{LLaVA-NeXT} \cite{liu2024llavanext}, that combine CLIP \cite{radford2021learning} as a visual encoder with Vicuna \cite{chiang2023vicuna} language decoder; \textit{Fuyu-8B} \cite{fuyu-8b}, an efficient model that integrates image patches directly into the transformer without an image encoder; \textit{Bunny} \cite{he2024bunny}, a versatile model family offering various combinations of vision encoders and LLM backbones; \textit{InstructBLIP} \cite{dai2024instructblip} builds on BLIP-2 \cite{li2023blip} for visual instruction tuning; \textit{LLaMA-Adapter} \cite{gao2023llamaadapterv2}, a parameter-efficient visual instruction model; and \textit{MiniGPT-4} \cite{zhu2023minigpt}, which aligns BLIP-2 with Vicuna via a projection layer.

\noindent
\textbf{Closed-source models. }We evaluated several closed-source multimodal models: \textit{GPT-4} \cite{openai2024gpt4}, which includes GPT-4V and GPT-4o variants, both pre-trained on large datasets and fine-tuned with RLHF \cite{christiano2017deep}; \textit{Claude 3 Opus} \cite{claude}, demonstrating competitive performance; and \textit{Gemini 1.5 Pro} \cite{geminiteam2024gemini} noted for surpassing GPT-4V across multiple benchmarks.

\subsection{Human Evaluators}
\label{subsec:human}
We conducted a preliminary human evaluation on our benchmark datasets, involving three evaluators who assessed $100$ uniformly sampled data points from each sub-category. Evaluators were shown one image and prompt at a time, with tasks including multiple-choice and true/false questions focused on depth and height ordering (example illustrated in Figure \ref{fig:prompt}). Their responses were compared to ground truth to calculate accuracy scores, establishing a baseline for human performance.

\subsection{Evaluation Metrics}
\label{subsec:metrics}
We evaluate our benchmark on the task of visual question answering (VQA), with accuracy being the performance metric on MCQ and True/False type questions. Evaluation is done across query attributes and scene density for probing the VLMs' depth and height perception. 

\subsection{Implementation Details}
\label{subsec:implementation}
All models are used in accordance to the provided evaluation code and model weights. The closed-source models were accessed through APIs provided through a paywall by the corresponding developing teams. For MCQ, the order of the given options are randomly generated, and ground truth is always randomly placed in one of those options. We have implemented already established practices ~\citep{liu2024mathbench, suzgun2022challenging} for creating options in multiple choice questions, randomizing both the position and the quantity of these options (up to 120 choices), and ensuring variability in the correct answer’s location. For the True/False questions, the ground truth is randomly selected between True and False.

\begin{figure*}
    \centering
    \includegraphics[width=\linewidth]{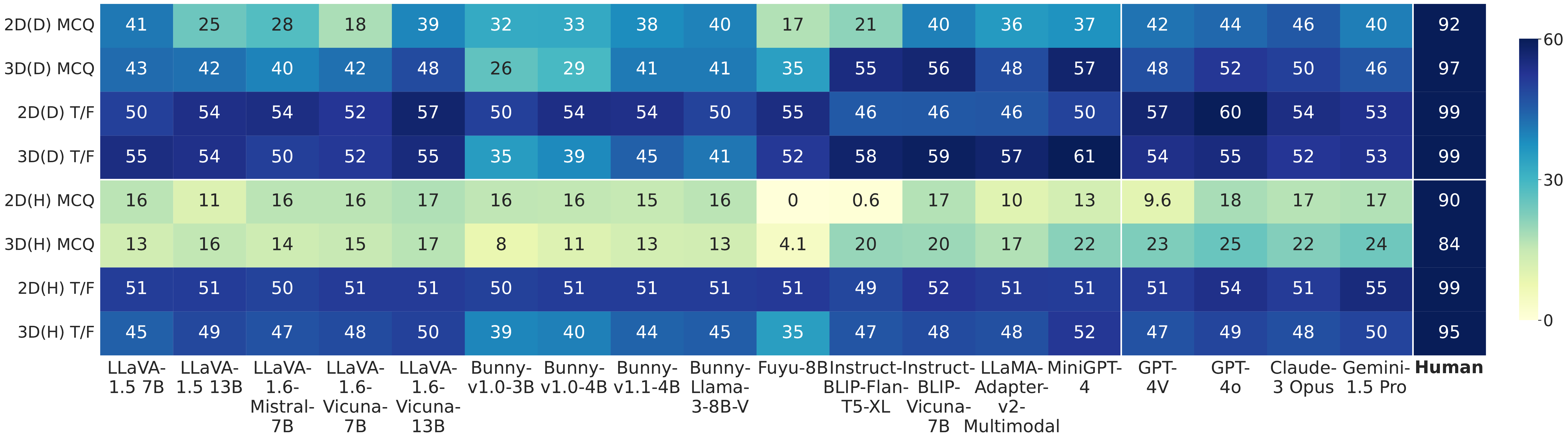}
    \caption{\textbf{Depth and height perception performance} on the proposed GeoMeter-2D and GeoMeter-3D dataset on MCQ and True/False (T/F) questions. D and H respectively denote depth, height performance. For example, 2D(D) MCQ and 2D(H) MCQ corresponds to respectively GeoMeter-2D depth and height performance on MCQ questions. Y-axis denotes the average performance across shape and query attributes and X-axis denotes all the evaluated models. Darker color denotes better performance.
    }
    \label{fig:main_heatmap}
\end{figure*}

\begin{figure*}[t!]
    \centering
    \includegraphics[width=.28\linewidth, height=3.6cm]{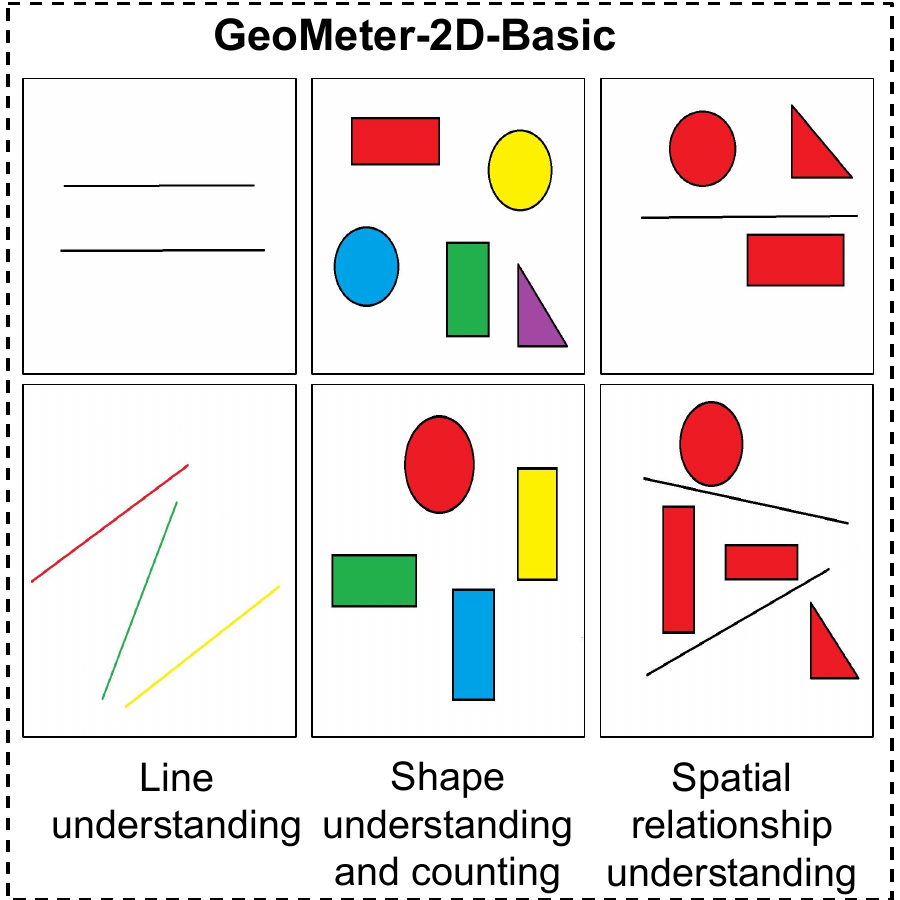}
    \hfill
    \includegraphics[width=.7\linewidth, height=3.5cm]{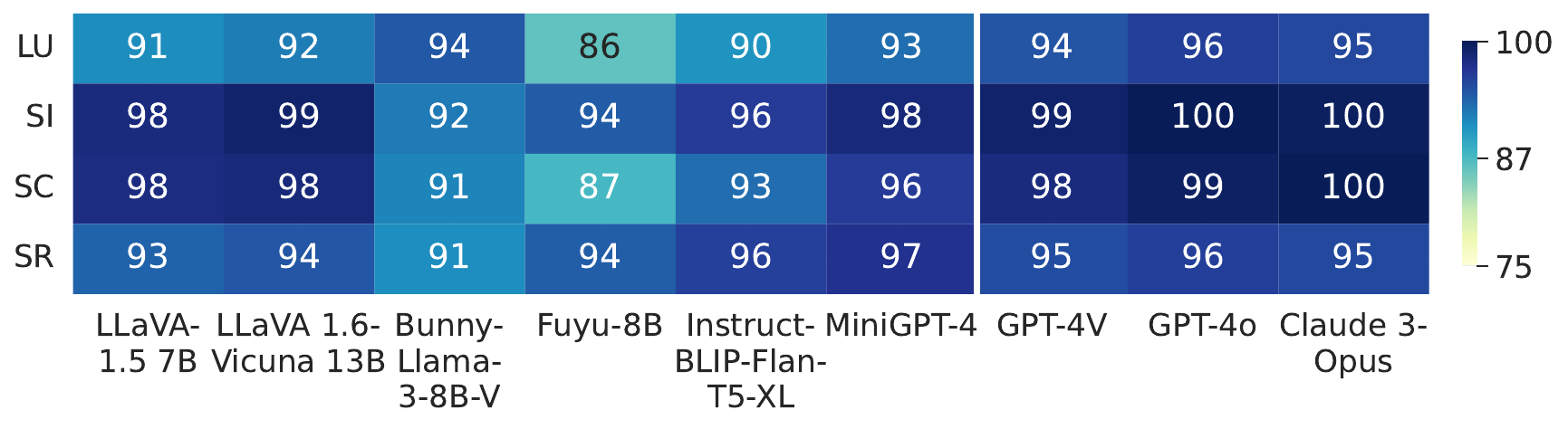}
    \caption{\textbf{Model behavior on basic understanding of shapes and size} on our created GeoMeter-2D-Basic dataset (samples on the \textit{left}). Performance of selected models on this dataset is shown in \textit{right}. Here, LU, SI, SC and SR respectively denote line understanding, shape identification, shape counting and spatial reasoning. Y-axis denotes performance accuracy of different categories and X-axis denotes evaluated models. Darker color denotes better performance.}
    \label{fig:basic}
    \vspace{-5mm}
\end{figure*}

\subsection{Results}
\label{subsec:results}
The performance of the selected models and human evaluators on the VQA task for MCQ and True/False type questions on the proposed benchmark datasets are shown in Table \ref{tab:avg}, where each model's performance represents the average accuracy of depth and height perception across all different query attributes and scene density. Depth and height category wise results are presented in Figure \ref{fig:main_heatmap}. Additional results across all query attributes and scene density are reported in the supplementary.

\section{Analysis and Discussion}
\label{sec:analysis}
\subsection{Model Behavior Analysis}

\textbf{Human evaluations confirm tasks are straightforward.} Despite the seemingly straightforward nature of depth and height perception tasks for humans, current Vision Language Models (VLMs) struggle to achieve comparable performance. Our initial human evaluations on our datasets show consistently high accuracy in both depth and height perception tasks (Table \ref{tab:avg}, Figure \ref{fig:main_heatmap}), demonstrating that humans can effortlessly solve these tasks. In contrast, VLMs exhibit significant limitations. This performance discrepancy highlights that while these tasks may appear trivial from a human perspective, they pose substantial challenges for foundation models. Moreover, the human evaluation serves as a baseline, indicating that these tasks should be within the capability of an advanced AI system. This clear gap in model performance underscores critical limitations in VLMs' visual reasoning, revealing that the models are not yet equipped to handle even elementary geometric understanding without additional sensory input. 

\begin{figure}
    \centering
    \includegraphics[width=\linewidth]{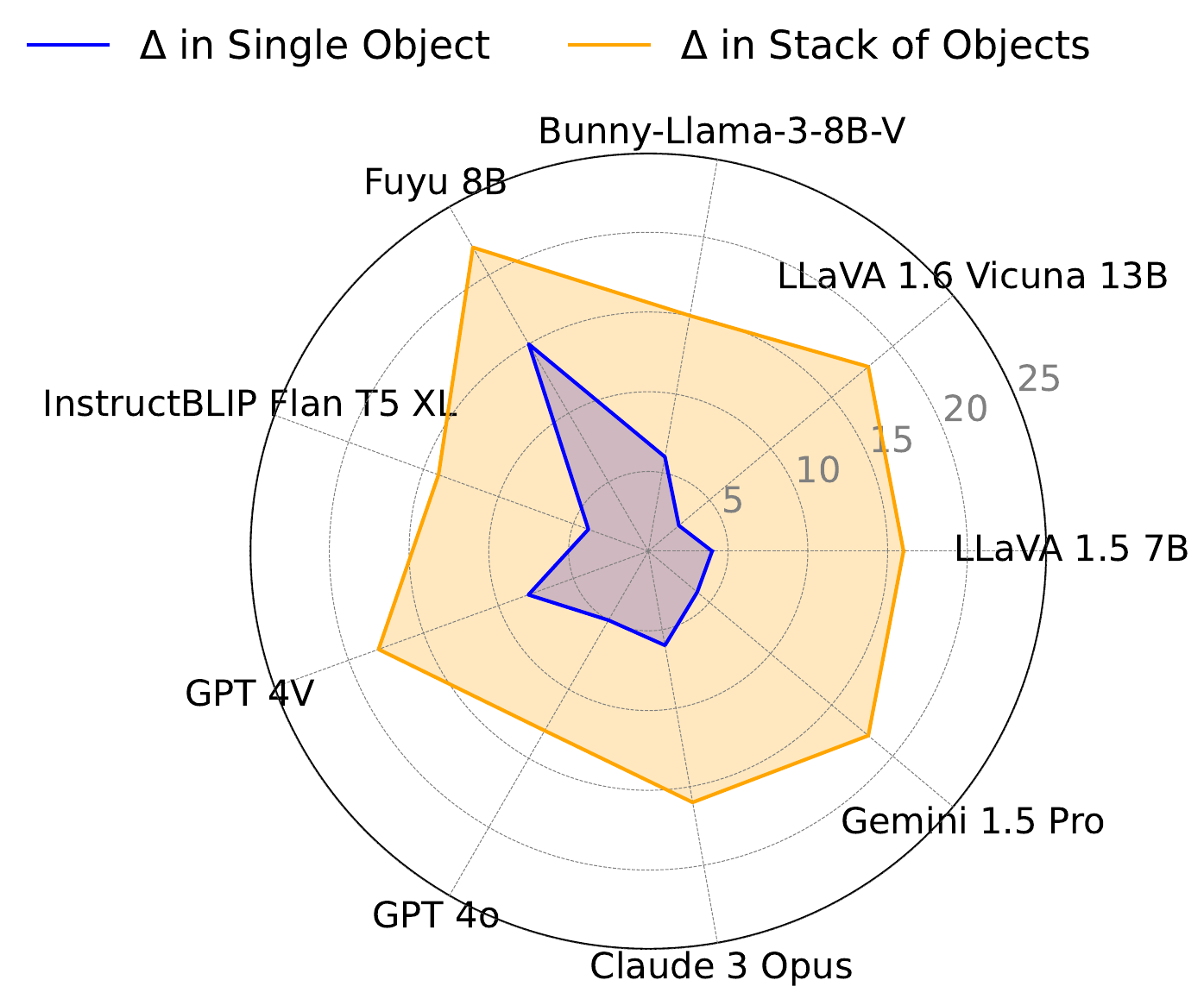}
    \caption{\textbf{Height perception is more challenging in stacked object arrangements than depth.} Here, $\Delta$ denotes performance gap between depth and height perception, which grows even larger with stacked arrangement of objects, as opposed to single objects. This suggests that while models struggle with height perception in general, stacked objects further degrade their performance.
    }
    \label{fig:depth_vs_height}
     \vspace{-10pt}   
\end{figure}
\noindent
\textbf{Models show basic visual reasoning capability but struggles in advance perception tasks.}  
We developed a specialized dataset called \textit{GeoMeter-2D-Basic} containing 30 image-text pairs (some samples shown in Figure \ref{fig:basic} \textit{left}) to evaluate the fundamental visual reasoning capabilities of Vision Language Models (VLMs). This dataset focuses on basic geometric tasks like  line understanding, shape recognition, shape counting, and assessing spatial relationships between shapes. The initial assessments using MCQs demonstrate high performance by models on these basic tasks, as detailed in Figure \ref{fig:basic} \textit{right}. Despite this proficiency in simple visual properties, results from Figure \ref{fig:main_heatmap} highlight that these same models struggle significantly with depth and height perception tasks involving similar shapes. This discrepancy underscores the benchmark's value in identifying gaps in VLMs' capabilities to handle more complex spatial reasoning, beyond mere shape recognition.

\noindent
\textbf{Height perception poses greater challenges than depth perception, especially in stacked object arrangements.} The superior performance of models in depth perception tasks, as compared to height perception (Figure \ref{fig:main_heatmap} row 1,2 vs row 5,6) is likely due to the availability of simpler depth cues—such as occlusion and perspective—in training datasets, which are relatively easy for VLMs to interpret. In contrast, we hypothesize height perception is more complex, requiring analysis of vertical object placement and relationships between object sizes in stacked arrangements. To further support our hypothesis, our analysis of single and stacked objects from the GeoMeter-3D dataset reveals (Figure \ref{fig:depth_vs_height}) that while the performance gap between depth and height tasks is minor for single objects, there is a substantial decline in performance for height tasks with stacked objects. This pattern suggests that height perception, especially with multiple objects stacked vertically, poses a greater challenge for VLMs than depth perception.

\noindent
\textbf{Models' limitation is due to inherent reasoning capability and not insufficient prompt detail.} To provide models with additional contextual information regarding visual cues with the help of intermediate reasoning, we implemented chain-of-thought prompting following ~\citep{wei2023chainofthought}. Chain of thought prompting enhances problem-solving by guiding models through logical reasoning steps, similar to human cognitive processes. 
\begin{figure}
    \centering
     \includegraphics[width=\linewidth, height=3.5cm]{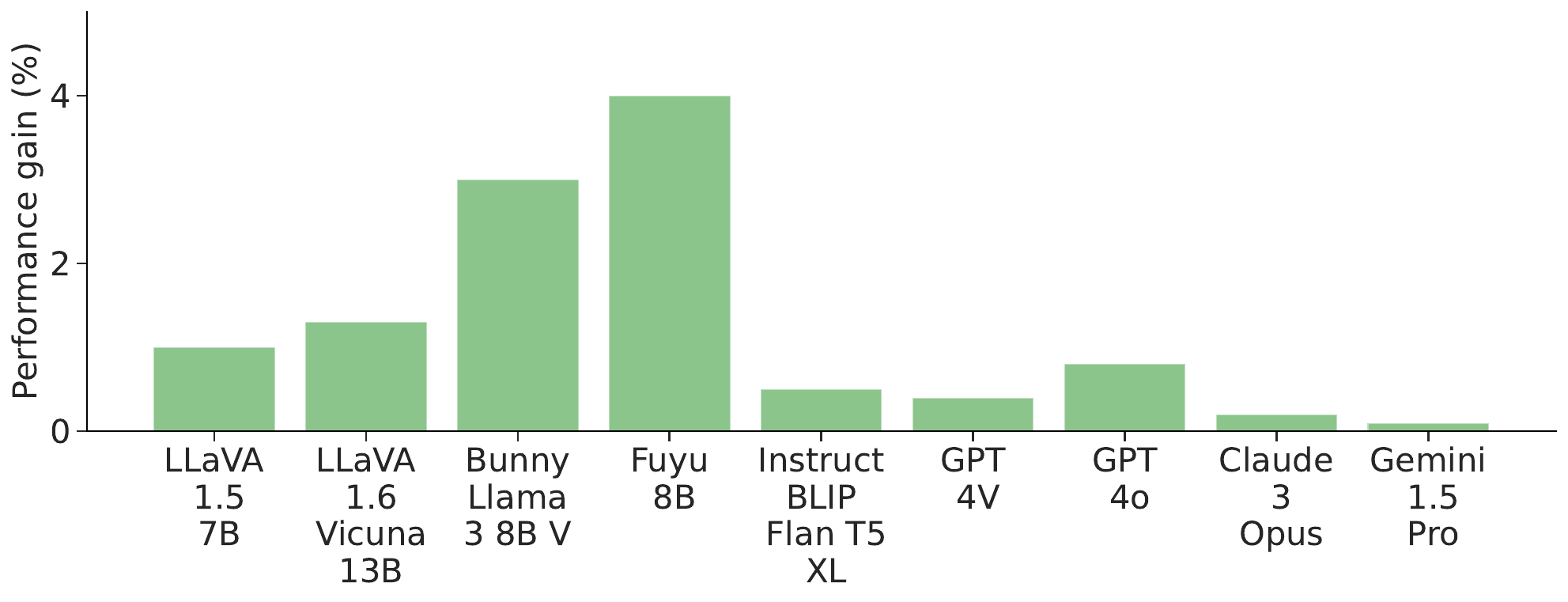}
    \caption{\textbf{Performance gain with chain of thought prompting over standard prompting} on subset of GeoMeter-3D dataset.}
    \label{fig:cot_result}
    \vspace{-10pt}
\end{figure}

\begin{figure*}[t!]
    \centering
    \includegraphics[width=\linewidth]{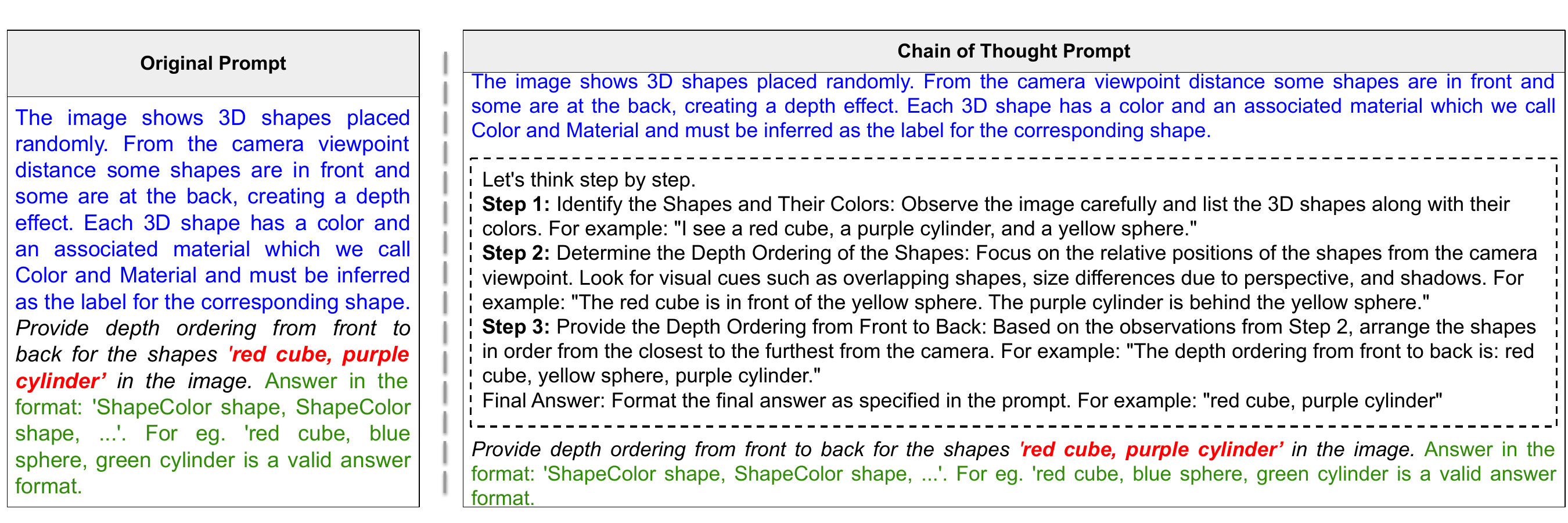}
    \caption{\textbf{Prompt engineering using chain of thought prompting.} Here the intermediate reasoning steps introduced in the engineered prompts of the GeoMeter-3D dataset is denoted by a dashed box.}
    \label{fig:cot}
\end{figure*}
\begin{figure*}
    \centering
    \includegraphics[width=.56\linewidth]{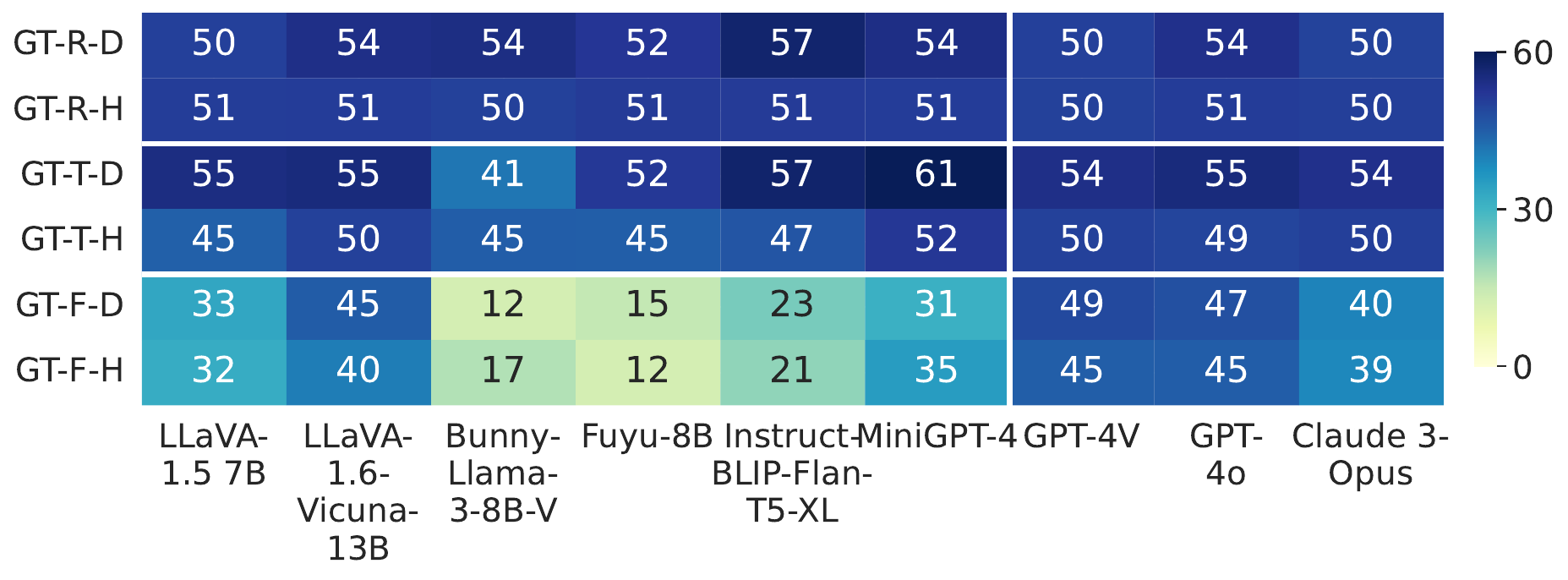}
    \includegraphics[width=.43\linewidth, height=3.5cm]{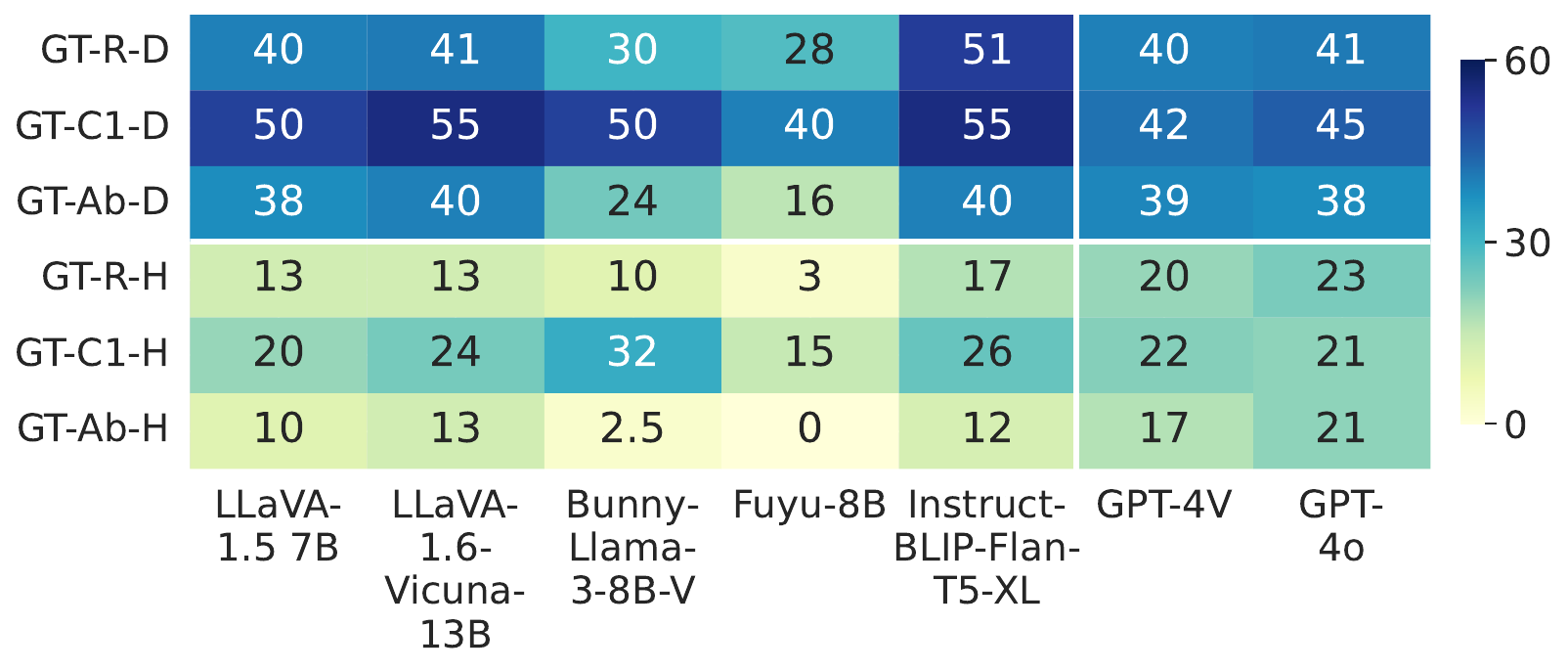}
    \caption{\textbf{Model bias analysis.} \textit{Left:} Effect of ground truth value in True/False questions. GT-R denotes randomly set ground truth between true and false; whereas GT-T/F denotes ground truth always true or always false. \textit{Right:} Effect of ground truth ordering in choices of MCQs. GT-C1 and GT-Ab denotes ground truth being choice 1 and not present respectively. The Y-axis denotes the average performance and X-axis denotes all the evaluated models. Darker colors denote better performance.}
    \label{fig:bias_analysis}
    \vspace{-4mm}
\end{figure*}

To evaluate its effectiveness, we selected a subset (100 image-text pairs) from the GeoMeter-3D dataset's depth category and created chain-of-thought prompts with intermediate reasoning steps. Testing top-performing models with these prompts showed only slight performance improvements (Figure \ref{fig:cot_result}), despite the highly detailed nature of these prompts (Figure \ref{fig:cot}). This marginal improvement suggesting that the models may already perform some internal reasoning with standard prompts. The findings indicate that limited depth and height perception performance is due to inherent model limitations in spatial understanding, underscoring the need for architectural advancements rather than solely relying on prompt engineering.

\subsection{Model Bias Analysis}
\label{subsec:prompt}
We conducted further analysis on the type of prompts to study any inherent biases in the models could be influencing their performance on MCQ and True/False type questions on a smaller subset (1600 image-text pairs uniformly selected from the depth and height categories) of the GeoMeter-3D dataset.

\noindent
\textbf{Some open-source models are more biased towards picking True over False than others.} 
The performance of some open-source models on True/False questions tends to hover around 50\% (Table \ref{tab:avg}), suggesting they might not be effectively distinguishing between true and false statements, potentially defaulting to random guesses. This is highlighted by experiments showing similar outcomes (Figure \ref{fig:bias_analysis} \textit{left}) when ground truth is random versus always set to "True," and a significant performance decline when it is always "False," indicating a bias towards predicting "True." This bias toward "True" may arise from imbalances in training data, where models are overexposed to affirmative statements or lack sufficient counterexamples of false statements. As a result, rather than demonstrating genuine understanding, these models often rely on heuristic patterns or shortcuts. Furthermore, this behavior highlights a deeper issue: the models' inability to engage in more nuanced decision-making or reasoning under uncertainty. True/False questions, though simple in format, test models' grasp of logical consistency and factual correctness - an area where many open-source models falter. By exposing such tendencies, this evaluation method provides valuable insight into where these models need refinement, particularly in developing the capacity for more context-driven predictions.

\noindent
\textbf{Some open source models are more biased towards picking the first choice in case of MCQ.} Experiments reveal that while closed-source models show consistent performance across various MCQ ground truth placements, open-source models exhibit a significant bias toward selecting the first option, particularly when the ground truth is positioned as the first choice (Figure \ref{fig:bias_analysis} \textit{right}). This bias could stem from the way training data is structured, where the first choice is frequently correct or if the models encounter more examples with answers listed early in the sequence, leading models to develop a preference for selecting it. Their performance drops notably when the correct answer is absent, suggesting these models struggle with identifying ``None of the above" options and may rely on heuristics rather than actual reasoning, leading to random selections. This reflects a limitation in their reasoning abilities, as they likely rely on pattern recognition rather than genuine understanding of the question and its context, which suggests that open-source models may lack sophisticated decision-making processes, opting for shortcuts when faced with challenging questions. 

\vspace{-2mm}
\section{Conclusion}
\label{sec:conclusion}
Our study highlights significant challenges in the depth and height reasoning capabilities of current Vision Language Models (VLMs). While these models demonstrate basic geometric understanding and spatial reasoning, they consistently struggle with more complex visual tasks, particularly depth and height perception, which remains underdeveloped. These shortcomings are not resolved by improved prompting alone, indicating an intrinsic limitation in the models' visual reasoning abilities. While our benchmark provides significant insights into the limitations of VLMs' perception abilities, it remains limited in scope. Expanding this benchmark to incorporate broader geometric reasoning, temporal dynamics, and higher-order reasoning tasks could offer a more comprehensive evaluation. Future work should focus on developing more targeted training strategies and benchmarks that address these perceptual weaknesses.

\section{Acknowledgement}
This research has benefitted from the Microsoft Accelerating Foundation Models Research (AFMR) grant program.

{
    \small
    \bibliographystyle{ieeenat_fullname}
    \bibliography{main}
}

\clearpage
\setcounter{page}{1}
\maketitlesupplementary

The supplementary will provide additional results and analysis on our proposed datasets. Additional results for GeoMeter 2D and GeoMeter 3D datasets are in Section \ref{sec_supp:quant} and Section \ref{sec_supp:qual}. Sections \ref{supp:impact}, \ref{supp:resources} respectively contain the broader impact and computational resources needed for our work.

\section{Additional Results}
\subsection{Quantitative Evaluation}
\label{sec_supp:quant}
Table \ref{tab:2d_depth}, Table \ref{tab:2d_height} present detailed results for the GeoMeter 2D dataset; and Table \ref{tab:3d_depth}, Table \ref{tab:3d_height} present detailed results for the GeoMeter 3D dataset. All of these results examine the impact of scene complexity (3 shapes vs 5 shapes), query attributes (color, labels), and question types (MCQ and True/False) on depth and height perception (respectively). While the main paper reports average results, the individual category-specific outcomes offer deeper insights. For instance, performance deteriorates with increased scene complexity (5 shapes) for many open-source models, highlighting the superior robustness of closed-source models under these conditions. Additionally, changes in query attributes show minimal impact on performance for most models, indicating their resilience to variations in query types. 

\begin{table*}[h!]
    \centering
     \caption{\textbf{Performance of the studied models on proposed GeoMeter-2D depth category.} Evaluation is done on the VQA task on MCQ and True/False type questions. Color, RL, PL are the query attributes. Here, RL, PL respectively denotes random numeric label, patterned numeric label. }

    \resizebox{\linewidth}{!}{
    \begin{tabular}{c|c|ccc|ccc|ccc|ccc}
        \hline
         &\multirow{2}{*}{Model}&\multicolumn{6}{c|}{Depth-3 shapes}&\multicolumn{6}{c}{Depth-5 shapes}\\
         &&\multicolumn{3}{c|}{MCQ}&\multicolumn{3}{c|}{T/F}&\multicolumn{3}{c|}{MCQ}&\multicolumn{3}{c}{T/F}\\
         &&Color&RL&PL&Color&RL&PL&Color&RL&PL&Color&RL&PL\\
         \hline
        \multirow{14}{*}{\rotatebox[]{90}{Open}}& LLaVA 1.5 7B & 48.0 & 37.5 & 54.5 & 49.0 & 54.5 & 47.0 & 36.5 & 31.0 &	39.0 & 45.0 & 56.0 & 49.5\\
         &LLaVA 1.5 13B&36.5&21.0&29.0&52.0&57.0&54.0&35.5& 15.0 &	11.0 & 54.5&	53.0&	54.0\\
         &LLaVA 1.6 Mistral 7B &44.0&34.5&25.0&55.5&54.5&52.5&28.5&	24.0&	11.0&	54.0&	56.0&	54.0\\
         &LLaVA 1.6 Vicuna 7B&37.0&20.5&13.0&54.5&50.5&49.5&29.0&7.0&1.0&50.5&	52.5&	55.0\\ 
         &LLaVA 1.6 Vicuna 13B&35.0&42.0&62.0&45.5&53.5&72.0&28.0&35.5&32.0&56.0&	54.0&	62.5\\
         &Bunny-v1.0-3B&41.5&	40.5&	38.5&	48.0&	45.5&	54.0&31.0&30.0&	13.5&	46.5&	52.5&	55.0\\
         &Bunny-v1.0-4B&38.0&	47.0&	33.5&	55.5&	55.5&	55.5&26.5&	29.5&	22.5&	52.5&	53.0&	53.0\\
         &Bunny-v1.1-4B&45.5&	47.5&	33.5&	52.5&	55.5&	55.5&34.0&	36.0&	31.5&	52.5&	53.0&	53.0\\
         &Bunny-Llama-3-8B-V&34.5&	45.0&	46.0&	41.0&	58.5&	51.5&27.5&36.5&	48.0&	48.5&	53.5&	46.0\\
         &Fuyu-8B&33.5& 17.0 &	4.5 & 58.5 & 55.5 & 55.5 & 30.0 & 	15.5& 3.0& 53.5& 53.0&	53.0\\
         &InstructBLIP-Flan-T5-XL&45.5& 8.5 & 0.0 & 44.5 & 44.5 & 44.5 & 32.0	& 40.0	& 0.0 &	47.0 & 47.0	& 47.0\\
         &InstructBLIP-Vicuna-7B&43.5 & 40.0 & 59.0&	49.5 & 44.0	& 43.0 & 32.0 & 31.0& 34.0	& 46.5 & 47.5 & 46.0\\
         &LLaMA-Adapter-v2-Multimodal&41.0&	40.0&	39.5&48.5&	45.5&	45.5&31.0&	30.0&	33.0&47&	45.5&	45.5\\
         &MiniGPT-4&42.0&	41.5&	43.0&52.0&	51.5&	51.5&34.0&	32.0&	30.0&48.5&	47.5&	47.5\\
         \hline
         \hline
         \multirow{3}{*}{\rotatebox[]{90}{Closed}}
         &GPT-4V& 45.0 & 49.0 & 41.5 & 54.5 & 57.0 & 61.5& 38.5 & 37.0 & 40.5	& 56.0 & 58.5 & 53.0\\
         &GPT-4o&47.5 & 44.5 & 47.0& 55.5 & 58.5 & 70.5 & 49.5	& 36.5 & 36.0 & 62.0 & 59.0 & 52.0\\
         &Claude 3 Opus&47.5&	40.5&	50&	51.5&	51.5&	56.5 & 36.5	& 36.0 & 41.0 & 52.5 & 51.5 & 56.0\\
         \hline
    \end{tabular}
    }
    \label{tab:2d_depth}
\end{table*}

\begin{table*}[t!]
    \centering
    \caption{\textbf{Performance of the studied models on proposed GeoMeter-2D height category.} Evaluation is done on the VQA task on MCQ and True/False type questions. Color, Label are the query attributes. Here, SP, \textoverline{SP} respectively denote w/ step, and w/o step.} 
    \resizebox{\linewidth}{!}{
    \begin{tabular}{c|c|cc|cc|cc|cc}
         \hline
         &\multirow{3}{*}{Model}&\multicolumn{4}{c|}{Height-3 towers \textoverline{SP}}&\multicolumn{4}{c}{Height-3 towers SP}\\
         &&\multicolumn{2}{c}{MCQ}&\multicolumn{2}{c|}{T/F}&\multicolumn{2}{c}{MCQ}&\multicolumn{2}{c}{T/F}\\
         &&Color&Label&Color&Label&Color&Label&Color&Label\\
         \hline
        \multirow{14}{*}{\rotatebox[]{90}{Open}}& LLaVA 1.5 7B & 15.5	& 18.0 & 50.0 & 54.0 & 21.0 & 16.5	& 49.5 & 57.0\\
         &LLaVA 1.5 & 15.5 & 9.0	& 49.0 & 54.0 & 14.5 & 10.0 & 49.0 & 56.9\\
         &LLaVA 1.6 Mistral 7B & 16.0 & 17.0	& 50.5 & 55.5 & 14.0	& 15.5 & 49.5 & 53.0\\
         &LLaVA 1.6 Vicuna 7B&14.0 & 19.0 & 49.0	& 55.0 & 18.5 & 18.0 & 50.0 & 58.0\\ 
         &LLaVA 1.6 Vicuna 13B& 19.0 & 19.0 & 49.5 & 54.0 & 13.5& 20.5 & 49.5 & 57.0\\
         &Bunny-v1.0-3B&13.5	& 17.5 & 49.0 & 51.0 & 18.5 & 20.0	& 49.0 & 57.0\\
         &Bunny-v1.0-4B&18.0	& 16.5 & 49.0 & 54.0 & 16.0 & 12.5 & 49.0 & 57.0\\
         &Bunny-v1.1-4B&11.0 & 18.5 & 49.0 & 54.0 & 19.0 & 15.0 & 49.0 & 57.0\\
         &Bunny-Llama-3-8B-V&15.0 & 15.5 & 49.0 & 54.5 & 14.5 & 18.0 & 49.0 & 53.5\\
         &Fuyu-8B&0.0 & 0.0 & 45.5 & 55.0 & 0.0 & 0.0 & 53.5 & 55.0\\
         &InstructBLIP-Flan-T5-XL&0.5 & 0.5 & 51.0 & 46.0 & 0.0 & 0.5 & 51.0 & 43.0\\
         &InstructBLIP-Vicuna-7B&19.0 & 16.0& 52.0 & 54.0 & 21.0 & 20.5 & 52.5 & 57.0\\
         &LLaMA-Adapter-v2-Multimodal&11.0&	9.0&52.0&	50.0&	13.0&	10.0&53.0&50.0\\
         &MiniGPT-4&13.0&12.0&54.0&52.5&15.0&14.0&54.0&51.5\\
         \hline
         \multirow{3}{*}{\rotatebox[]{90}{Closed}}
         &GPT-4V&6.5	& 7.0 & 48.0 & 55.5 & 3.0 & 10.0 & 48.5 & 56.0\\
         &GPT-4o&21.0 & 17.0 & 57.0 & 53.0 & 17.5 & 15.5 & 51.5	& 56.5\\
         &Claude 3 Opus&15.0& 13.5 & 50.5 & 51.5& 16.0 & 18.5 & 50.0 & 56.0\\
         \hline
         \hline
         &\multirow{3}{*}{Model}&\multicolumn{4}{c|}{Height-5 towers \textoverline{SP}}&\multicolumn{4}{c}{Height-5 towers SP}\\
         &&\multicolumn{2}{c}{MCQ}&\multicolumn{2}{c|}{T/F}&\multicolumn{2}{c}{MCQ}&\multicolumn{2}{c}{T/F}\\
         &&Color&Label&Color&Label&Color&Label&Color&Label\\
         \hline
        \multirow{14}{*}{\rotatebox[]{90}{Open}}& LLaVA 1.5 7B & 14.0 & 14.0 & 46.0 & 47.0& 14.0 & 18.5 & 51.5 & 51.0\\
         &LLaVA 1.5 13B&12.0 & 9.0 & 52.0 & 49.0 & 8.5.0 & 8.0 & 49.0 & 48.0\\
         &LLaVA 1.6 Mistral 7B &16.0& 14.5 & 46.0 & 46.0 & 17.5 & 20.5 & 48.0 & 51.0\\
         &LLaVA 1.6 Vicuna 7B& 16.0 & 13.5 & 51.5 & 49.5 & 16.0 & 15.0 & 48.5 & 49.0\\ 
         &LLaVA 1.6 Vicuna 13B&16.5 & 16.0 & 52.0 & 49.0 & 20.0 & 14.5 & 49.0 & 49.0\\
         &Bunny-v1.0-3B&13.0 & 11.5 & 50.5 & 44.0 & 12.5 & 19.5 & 49.0 & 50.5\\
         &Bunny-v1.0-4B& 16.0& 14.5 & 52.0 & 49.0 & 14.0 & 17.0 & 49.0 & 49.0\\
         &Bunny-v1.1-4B&14.5 & 13.0 & 52.0 & 49.0 & 12.0 & 18.0 & 49.0 & 49.0\\
         &Bunny-Llama-3-8B-V&15.0& 15.0 & 52.0 & 47.5 & 14.5 & 21.0 & 49.0 & 49.5\\
         &Fuyu-8B&0.0 & 0.0 & 52.5 & 51.5 & 0.0 & 0.0 & 49.0 & 46.5\\
         &InstructBLIP-Flan-T5-XL& 0.0 & 1.5 & 48.0 & 51.0 & 0.0 & 1.5 & 51.0 & 51.0\\
         &InstructBLIP-Vicuna-7B& 15.0 & 11.0 & 52.5& 49.0 & 15.0 & 16.0 & 48.5 & 49.0\\
         &LLaMA-Adapter-v2-Multimodal&10.5&8.5&51.0&52&9.5&9.0&50.0&51.5\\
         &MiniGPT-4&13.5&10.0&52.0&50.0&12.0&10.5&51.0&49.5\\
         \hline
         \hline
         \multirow{3}{*}{\rotatebox[]{90}{Closed}}
         &GPT-4V& 17.5& 12.5 & 51.5 & 50.0 & 14.0 & 6.5 & 50.0 & 49.0\\
         &GPT-4o&18.0 & 18.5 & 59.5 & 50.0 & 19.0 & 19.0 & 51.0 & 52.0\\
         &Claude 3 Opus&19.5	& 14.0 & 48.5 & 51.5 & 13.0 & 19.5 & 47.5 & 48.5\\

         \hline
    \end{tabular}
    }
    \label{tab:2d_height}
\end{table*}

\begin{table*}[t!]
    \centering
    \caption{\textbf{Performance of the studied models on proposed GeoMeter-3D height category.}  Evaluation is done on the VQA task on MCQ and True/False type questions. Color, ColMat are the query attributes. Here, ColMat denotes color+material} 
    \resizebox{\linewidth}{!}{
    \begin{tabular}{c|c|cc|cc|cc|cc}
        \hline
         &\multirow{3}{*}{Model}&\multicolumn{4}{c|}{Depth-3 shapes}&\multicolumn{4}{c}{Depth-5 shapes}\\
         &&\multicolumn{2}{c|}{MCQ}&\multicolumn{2}{c|}{T/F}&\multicolumn{2}{c|}{MCQ}&\multicolumn{2}{c}{T/F}\\
         &&Color&ColMat&Color&ColMat&Color&ColMat&Color&ColMat\\
         \hline
        \multirow{14}{*}{\rotatebox[]{90}{Open}}& LLaVA 1.5 7B &49.1&42.5&59.4&53.8& 43.1&	37.5&55.7&	50.4\\
         &LLaVA 1.5 13B&51.3&45.9&61.9&58.4&37.3&35.1&50.3&44.3\\
         &LLaVA 1.6 Mistral 7B &47.1	&45.3&51.9&	50.6&34.8&30.8&50.3&48.9\\
         &LLaVA 1.6 Vicuna 7B&48.8&	47.3&
        61.9&	58.3&40.2&32.9&45.9&40.2\\ 
         &LLaVA 1.6 Vicuna 13B&51.8&	50.3&
        64.2&	61.2&48.3&42.9&50.2&45.9\\

         &Bunny-v1.0-3B&34.8&29.3&40.2&35.8&21.9&18.3&34.8&29.8\\
         &Bunny-v1.0-4B&34.2&30.8&45.3&43.2&28.2&23.2&34.9&30.7\\
         &Bunny-v1.1-4B&45.2&40.3&44.2&42.9&40.2&38.3&48.3&42.9\\
         &Bunny-Llama-3-8B-V&44.2&42.1&45.2&40.8&40.8&35.9&40.8&38.3\\
         &Fuyu-8B&41.8&38.4&59.3&	51.8&30.5&	27.5&48.3&47.2\\
         &InstructBLIP-Flan-T5-XL&58.3&	54.2&	55.3&	51.3&	61.9&	59.3&	54.9&	53.8\\
         &InstructBLIP-Vicuna-7B&57.4&	56.3&	56.9&	55.4&		60.2&	57.3&	59.9&	58.6\\
         &LLaMA-Adapter-v2-Multimodal&52.9&	48.3&	47.3&	44.2&	59.8&	56.8&	57.8&	54.7\\
         &MiniGPT-4&60.3&	56.3&	57.8&	54.8&65.3&	62.9&	60.3&	54.8\\
         \hline
         \hline
         \multirow{3}{*}{\rotatebox[]{90}{Closed}}
        &GPT-4V&54.3&50.1&63.9&60.2&45.3&40.9&48.4&43.2\\
         &GPT-4o&59.9&52.9&65.9&60.3&50.3&44.3&50.3&44.8\\
         &Claude 3 Opus&56.3&53.9&57.3&52.3&47.3&43.2&51.8&47.4\\
         
         \hline
    \end{tabular}
    }
    \label{tab:3d_depth}
\end{table*}

\begin{table*}[t!]
    \centering
    \caption{\textbf{Performance of the studied models on proposed GeoMeter-3D height category.}  Evaluation is done on the VQA task on MCQ and True/False type questions. Color, ColMat are the query attributes. Here, ColMat, SP, \textoverline{SP} respectively denotes color+material, w/ step, and w/o step.} 
    \resizebox{\linewidth}{!}{
    \begin{tabular}{c|c|cc|cc|cc|cc}
         \hline
         &\multirow{3}{*}{Model}&\multicolumn{4}{c|}{Height-3 towers \textoverline{SP}}&\multicolumn{4}{c}{Height-3 towers SP}\\
         &&\multicolumn{2}{c}{MCQ}&\multicolumn{2}{c|}{T/F}&\multicolumn{2}{c}{MCQ}&\multicolumn{2}{c}{T/F}\\
         &&Color&ColMat&Color&ColMat&Color&ColMat&Color&ColMat\\
         \hline
        \multirow{14}{*}{\rotatebox[]{90}{Open}}
        &LLaVA 1.5 7B                & 20.3 & 12.9 & 48.2 & 40.8 & 18.8 & 8.1  & 46.3 & 40.3 \\
        &LLaVA 1.5 13B               & 22.8 & 18.3 & 52.1 & 48.9 & 19.9 & 15.8 & 48.2 & 45.9 \\
        &LLaVA 1.6 Mistral 7B        & 21.9 & 18.7 & 49.9 & 42.7 & 18.3 & 12.8 & 47.9 & 44.3 \\
        &LLaVA 1.6 Vicuna 7B         & 20.8 & 18.9 & 48.7 & 44.8 & 18.7 & 12.7 & 49.7 & 43.8 \\
        &LLaVA 1.6 Vicuna 13B        & 24.9 & 19.8 & 50.7 & 47.3 & 20.8 & 17.3 & 50.2 & 45.9 \\
         
         &Bunny-v1.0-3B               & 12.4 & 9.4  & 51.4 & 50.4 & 9.4  & 5.3  & 42.9 & 40.3 \\
         &Bunny-v1.0-4B               & 14.9 & 10.4 & 51.8 & 48.3 & 12.9 & 10.5 & 44.3 & 41.7 \\
         &Bunny-v1.1-4B               & 15.9 & 12.7 & 54.8 & 52.6 & 13.7 & 11.8 & 50.3 & 48.5 \\
        &Bunny-Llama-3-8B-V          & 16.3 & 12.8 & 55.7 & 53.9 & 14.9 & 13.9 & 52.9 & 49.3 \\
        &Fuyu-8B                     & 9.3  & 7.9  & 40.2 & 35.4 & 5.9  & 3.9  & 37.9 & 34.7 \\
        &InstructBLIP-Flan-T5-XL     & 25.1 & 20.9 & 53.8 & 50.3 & 22.9 & 20.4 & 50.3 & 48.2 \\
        &InstructBLIP-Vicuna-7B      & 24.9 & 21.9 & 54.3 & 52.9 & 20.8 & 18.9 & 52.7 & 49.3 \\
        &LLaMA-Adapter-v2-Multimodal & 23.9 & 20.3 & 49.3 & 47.8 & 20.2 & 18.7 & 48.2 & 45.8 \\
        &MiniGPT-4                   & 26.9 & 24.8 & 54.8 & 53.7 & 24.8 & 20.4 & 53.8 & 51.8 \\
        \hline
        \multirow{3}{*}{\rotatebox[]{90}{Closed}}
        &GPT-4V                      & 28.8 & 25.9 & 48.3 & 48.0 & 27.1 & 26.9 & 46.0 & 43.9 \\
        &GPT-4o                      & 30.5 & 28.9 & 50.9 & 49.2 & 28.9 & 27.8 & 49.3 & 46.8 \\
        &Claude 3 Opus               & 28.3 & 24.0 & 51.8 & 48.3 & 26.1 & 22.0 & 47.3 & 43.0\\
         
         \hline
         \hline
         &\multirow{3}{*}{Model}&\multicolumn{4}{c|}{Height-5 towers \textoverline{SP}}&\multicolumn{4}{c}{Height-5 towers SP}\\
         &&\multicolumn{2}{c}{MCQ}&\multicolumn{2}{c|}{T/F}&\multicolumn{2}{c}{MCQ}&\multicolumn{2}{c}{T/F}\\
         &&Color&ColMat&Color&ColMat&Color&ColMat&Color&ColMat\\
         \hline
        \multirow{14}{*}{\rotatebox[]{90}{Open}}
& LLaVA 1.5 7B                & 12.9 & 10.4 & 48.3 & 42.3 & 10.4 & 9.3  & 47.3 & 43.8 \\
 & LLaVA 1.5 13B               & 13.9 & 11.3 & 50.3 & 49.2 & 11.8 & 10.5 & 49.3 & 47.3 \\
 & LLaVA 1.6 Mistral 7B        & 11.0 & 9.3  & 50.4 & 47.3 & 10.3 & 8.3  & 47.0 & 46.9 \\
 & LLaVA 1.6 Vicuna 7B         & 13.9 & 10.3 & 51.9 & 49.2 & 11.8 & 10.8 & 50.8 & 47.1 \\
 & LLaVA 1.6 Vicuna 13B        & 15.9 & 12.3 & 54.1 & 50.3 & 12.9 & 9.3  & 52.9 & 48.3 \\
 & Bunny-v1.0-3B               & 9.2  & 4.2  & 34.3 & 28.4 & 7.3  & 6.9  & 33.2 & 30.9 \\
 & Bunny-v1.0-4B               & 11.9 & 9.3  & 35.3 & 30.4 & 9.3  & 5.3  & 34.3 & 33.9 \\
 & Bunny-v1.1-4B               & 13.9 & 11.4 & 39.3 & 36.3 & 12.9 & 10.2 & 37.3 & 33.9 \\
 & Bunny-Llama-3-8B-V          & 13.3 & 12.1 & 38.3 & 37.9 & 10.3 & 9.9  & 36.3 & 35.9 \\
 & Fuyu-8B                     & 4.2  & 1.8  & 35.3 & 30.0 & 0.0  & 0.0  & 32.8 & 31.9 \\
 & InstructBLIP-Flan-T5-XL     & 19.8 & 18.9 & 47.2 & 42.1 & 16.3 & 15.9 & 42.9 & 38.3 \\
 & InstructBLIP-Vicuna-7B      & 18.3 & 17.9 & 46.3 & 45.8 & 17.0 & 16.9 & 43.9 & 42.7 \\
 & LLaMA-Adapter-v2-Multimodal & 15.3 & 12.8 & 48.3 & 48.0 & 13.9 & 12.8 & 47.4 & 45.4 \\
 & MiniGPT-4                   & 20.8 & 19.3 & 53.2 & 50.2 & 19.2 & 16.0 & 49.3 & 47.3 \\
 \hline
         \multirow{3}{*}{\rotatebox[]{90}{Closed}}
 & GPT-4V                      & 19.3 & 17.3 & 48.4 & 47.8 & 18.3 & 16.9 & 47.0 & 46.3 \\
 & GPT-4o                      & 22.6 & 21.9 & 51.9 & 50.3 & 20.9 & 19.6 & 49.4 & 47.4 \\
 & Claude 3 Opus               & 21.9 & 19.3 & 49.3 & 47.0 & 19.7 & 15.9 & 48.9 & 44.8\\

         \hline
    \end{tabular}
    }
    \label{tab:3d_height}
\end{table*}

\subsection{Qualitative Examples}
\label{sec_supp:qual}
\begin{figure*}[t!]
    \centering
    \includegraphics[width=\textwidth, height=13cm]{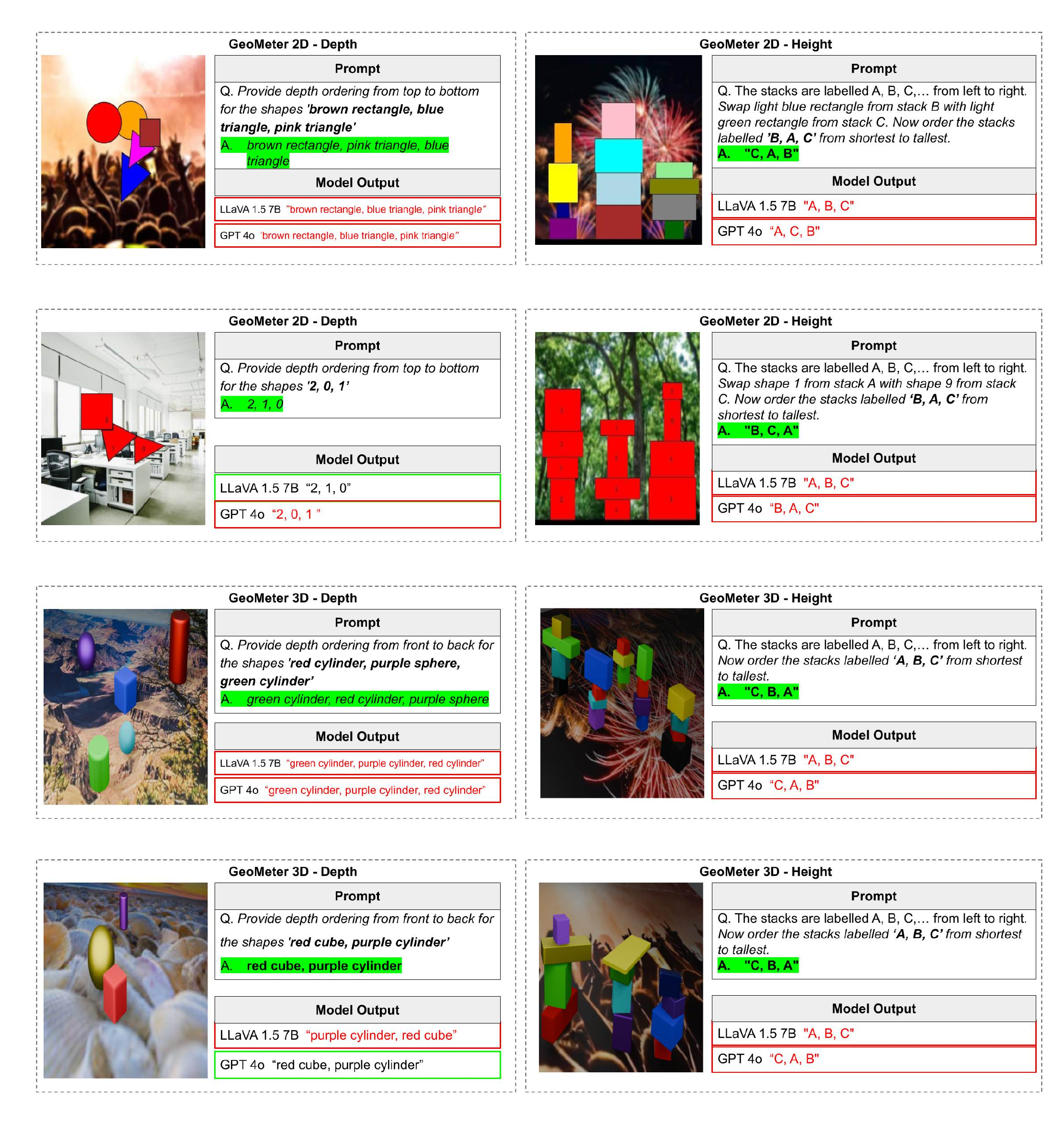}
    \caption{\textbf{Depth and height perception of open and closed models.} Here we show the prediction of LLaVA 1.5 7B and GPT 4o. Here Q and A respectively denote Question and Ground Truth Answer. Green and Red boxes respectively denote correct and incorrect prediction.}
    \label{fig:qual_example}
\end{figure*}

\begin{figure*}[t!]
    \centering
    \includegraphics[width=\textwidth, height=20cm]{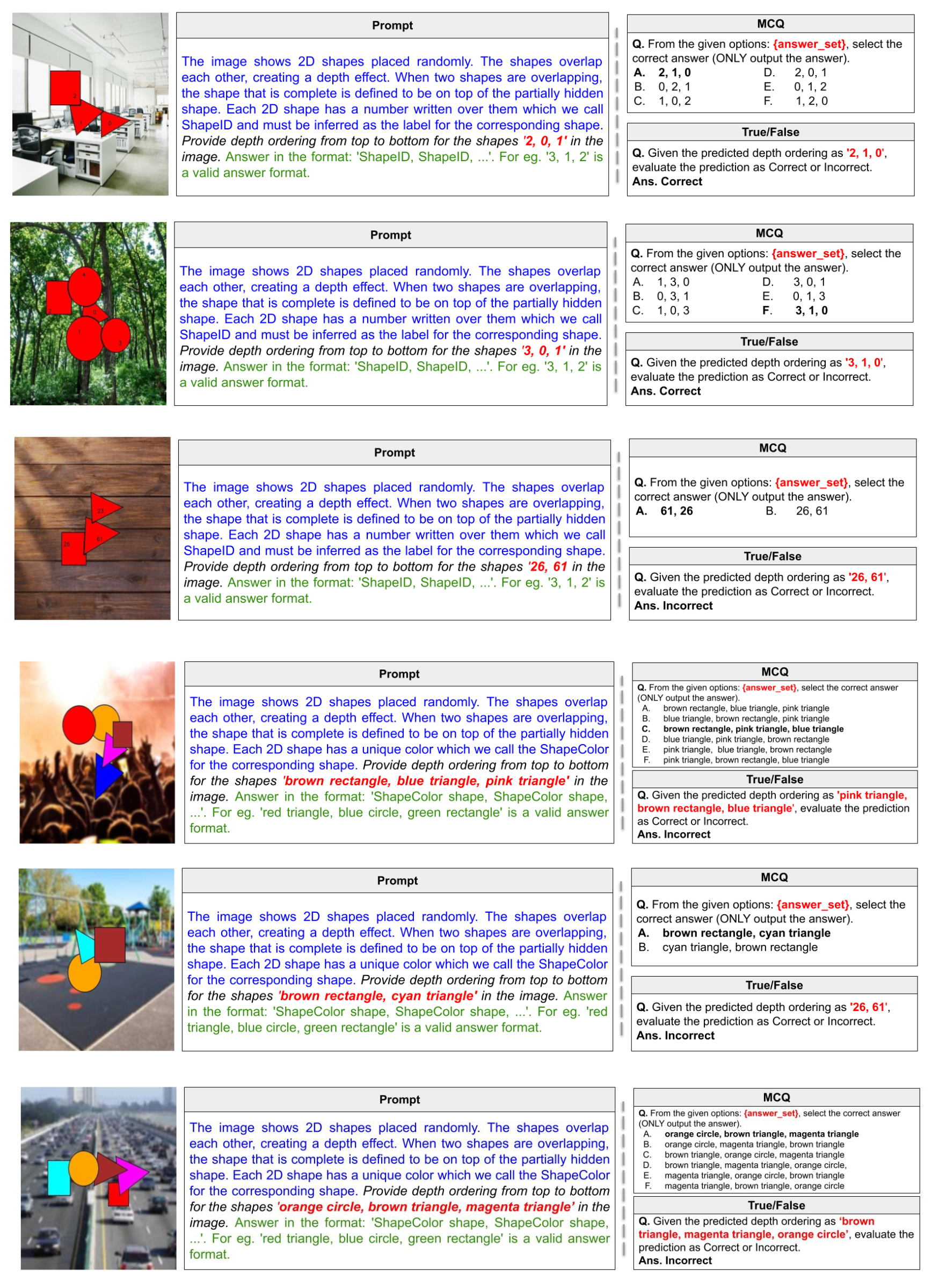}
    \caption{\textbf{Samples from GeoMeter-2D dataset - depth category.} Here each row represents one image and its corresponding prompt along with MCQ and True/False questions. First three rows show samples for labels as query attribute, whereas last three rows show samples for color as query attribute.}
    \label{fig:depth_sample_2d}
\end{figure*}

\begin{figure*}[t!]
    \centering
    \includegraphics[width=\textwidth, height=20cm]{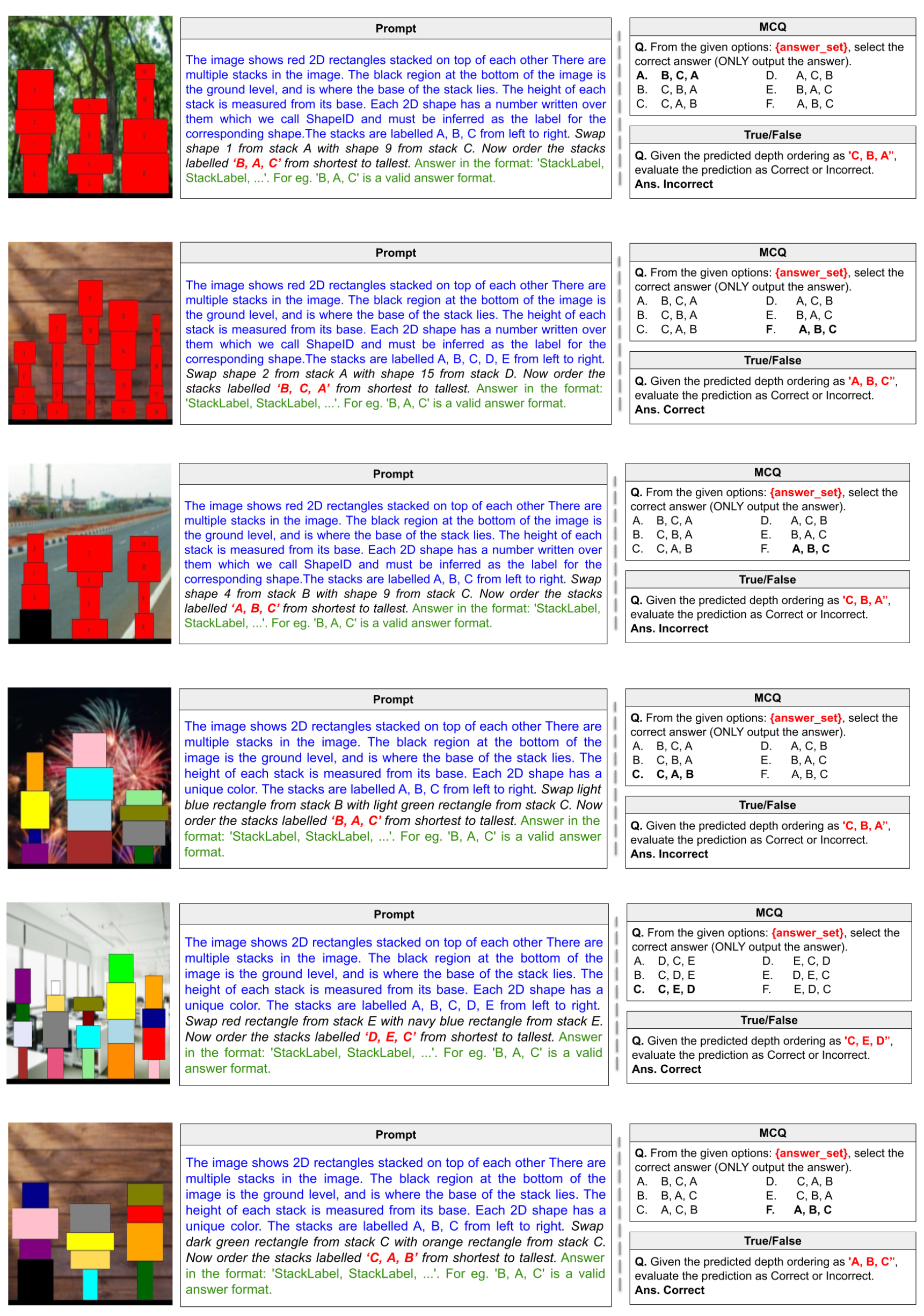}
    \caption{\textbf{Samples from GeoMeter-2D dataset - height category.} Here each row represents one image and its corresponding prompt along with MCQ and True/False questions. First three rows show samples for labels as query attribute, whereas last three rows show samples for color as query attribute}
    \label{fig:height_sample_2d}
\end{figure*}

\begin{figure*}[t!]
    \centering
    \includegraphics[width=\textwidth, height=20cm]{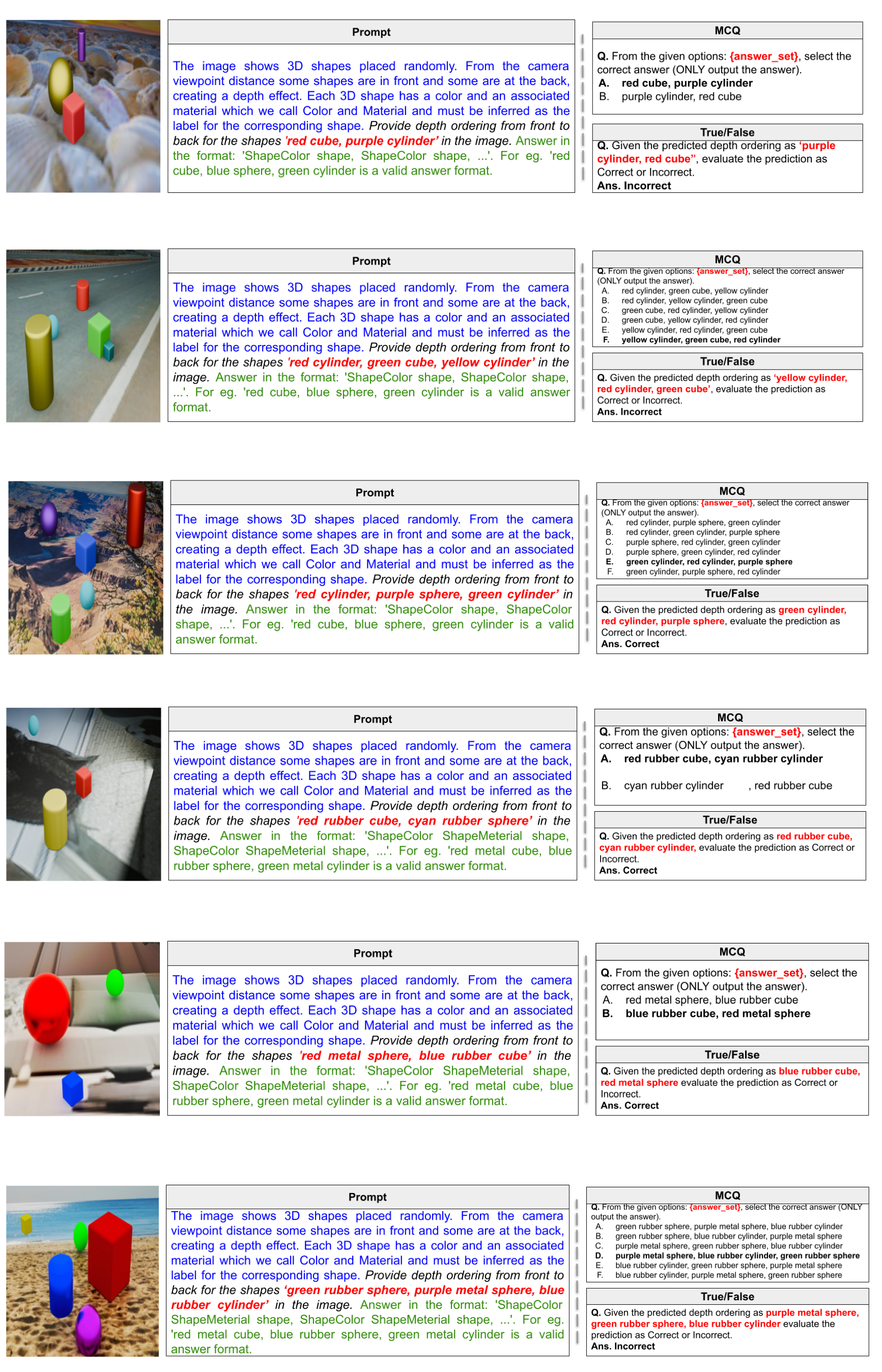}
    \caption{\textbf{Samples from GeoMeter-3D dataset - depth category.} Here each row represents one image and its corresponding prompt along with MCQ and True/False questions. First three rows show samples for color as query attribute, whereas last three rows show samples for color+material as query attribute}
    \label{fig:depth_sample_3d}
\end{figure*}

\begin{figure*}[t!]
    \centering
    \includegraphics[width=\textwidth, height=20cm]{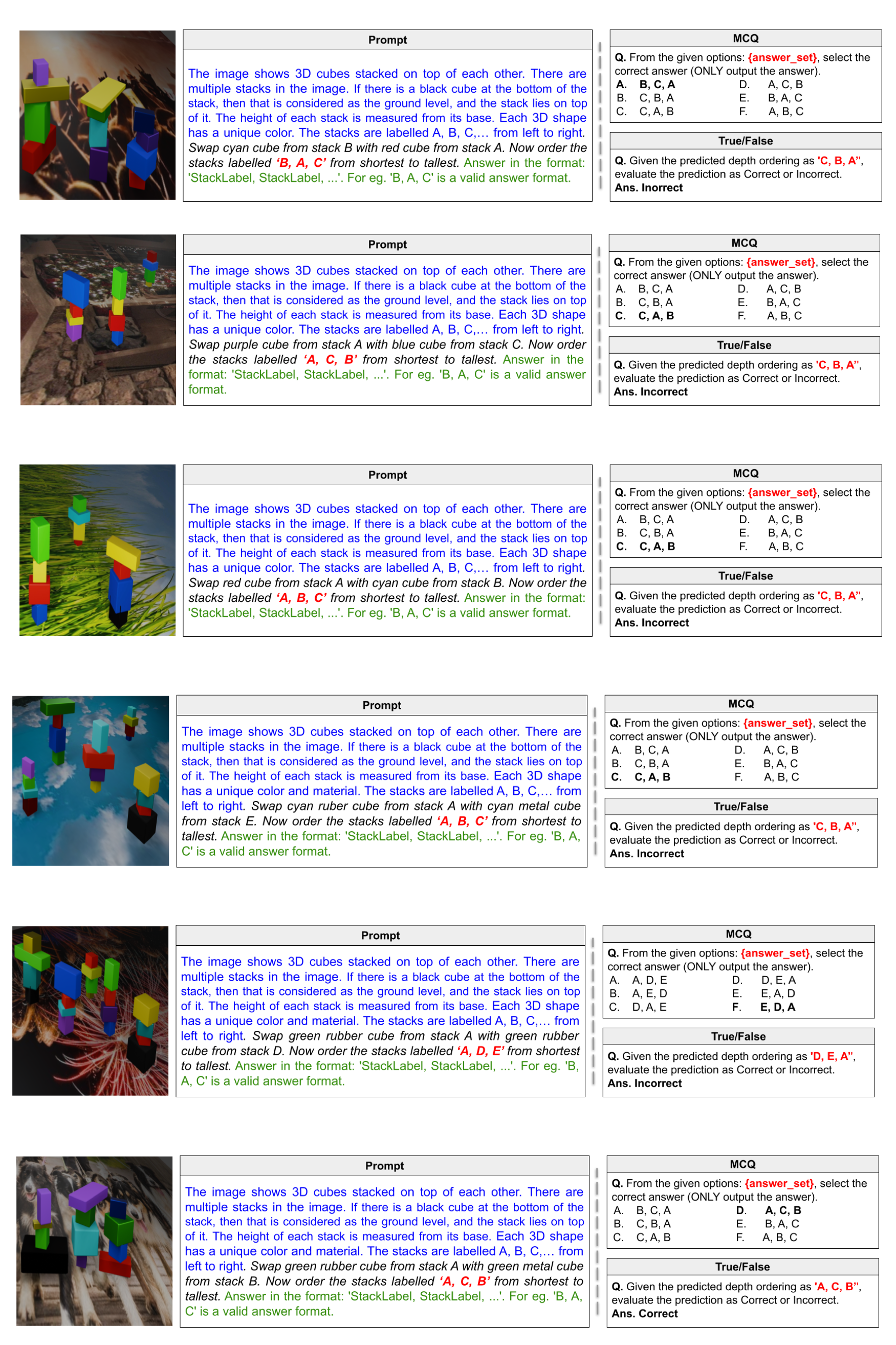}
    \caption{\textbf{Samples from GeoMeter-3D dataset - height category.} Here each row represents one image and its corresponding prompt along with MCQ and True/False questions. First three rows show samples for color as query attribute, whereas last three rows show samples for color+material as query attribute}
    \label{fig:height_sample_3d}
\end{figure*}

\begin{figure*}[t!]
    \centering
    \includegraphics[width=\linewidth, height=20cm]{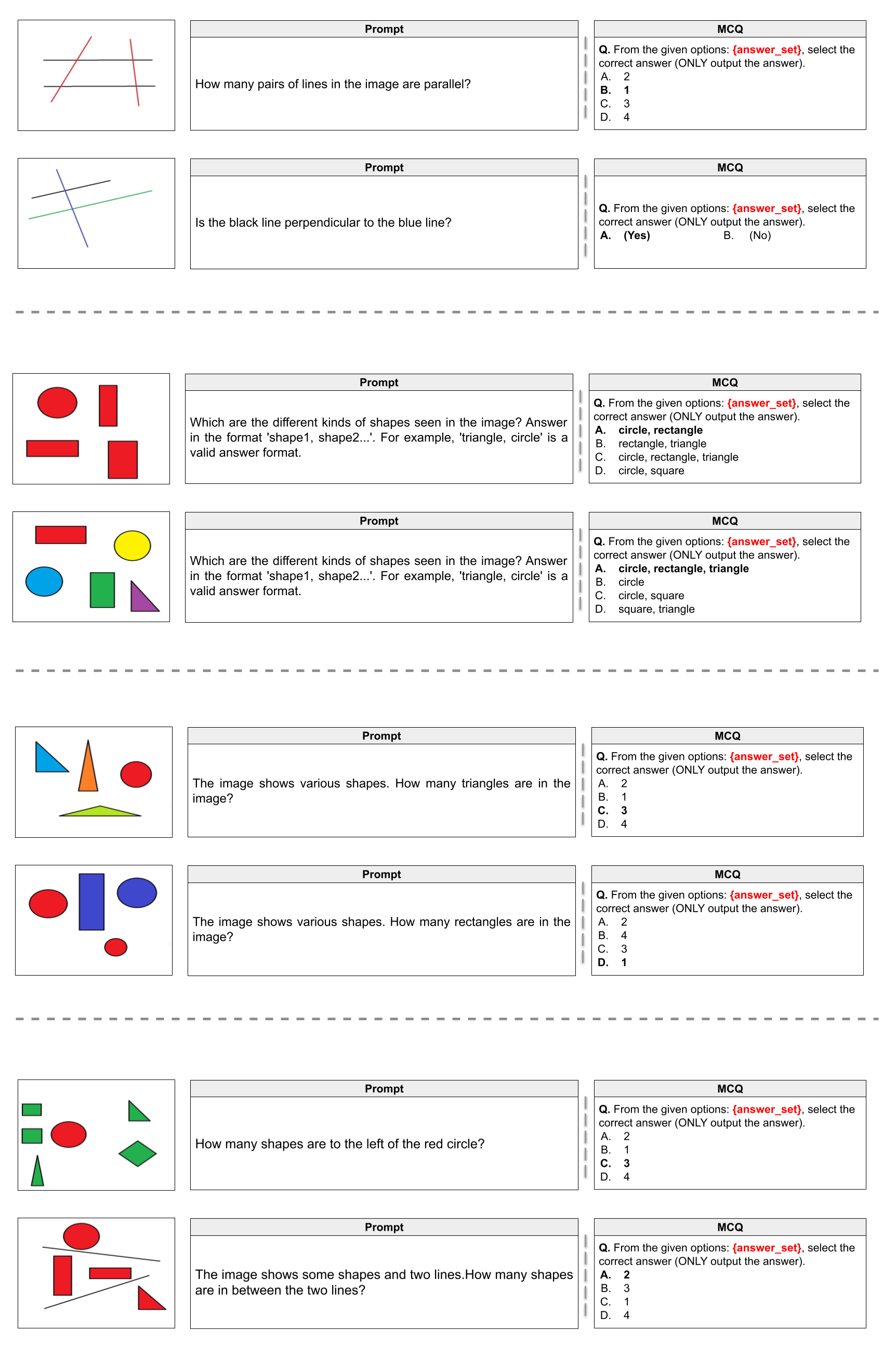}
    \caption{\textbf{Samples from GeoMeter-2D-Basic dataset.} Here each two rows respectively represent line understanding, shape identification, shape counting and spatial relationship categories. Each row shows one image and its corresponding prompt along with the MCQ.}
    \label{fig:sample_basic}
    \vspace{5mm}
\end{figure*}

Figure \ref{fig:qual_example} displays sample predictions from both open and closed models, highlighting their challenges with depth and height perception. The examples particularly emphasize the models' inaccuracies, especially in height perception, showcasing their limitations in spatial understanding. This figure includes predictions from the best-performing models in the open (LLaVA 1.5 7B) and closed (GPT 4o) categories.  Figures \ref{fig:depth_sample_2d} and \ref{fig:height_sample_2d} present examples from the GeoMeter-2D dataset, including the specific prompts for both MCQ and True/False questions, serving as visual aids for the evaluations discussed. Similarly, Figures \ref{fig:depth_sample_3d} and \ref{fig:height_sample_3d}, showcase samples and corresponding prompts from the GeoMeter-3D depth and height category, respectively. These figures provide insights into the different scenarios and questions used to assess depth and height perception across various data types. Additionally, Figure \ref{fig:sample_basic} features image-text pairs from the GeoMeter-2D Basic dataset, highlighting the initial stages of evaluating the models' capabilities in recognizing basic properties.

\section{Broader Impact}
\label{supp:impact}
To our understanding, there are no negative societal impacts of our work. The goal of this work was to evaluate the depth and height perception capabilities of models that may later be used in real-world settings. This research provides insights into the depth and height perception capabilities of vision language models (VLMs), significantly impacting practical applications like autonomous driving, augmented reality, and assistive technologies. This work not only advances theoretical understanding but also opens up new possibilities for real-world applications. 

\section{Computational Resources}
\label{supp:resources}
All experiments were run on an internal cluster. Each run used a single NVIDIA GPU, with memory ranging from 16GB-24GB. 

\end{document}